%% file: bmvc_final.tex
\documentclass{bmvc2k}
\usepackage{amssymb}
\usepackage{amsfonts}
\usepackage{amsmath}
\usepackage{algorithm}
\usepackage{algpseudocode}
\usepackage{tikz}
\usepackage{graphicx}
\usetikzlibrary{arrows.meta,positioning,fit}
\usepackage{xcolor}
\usepackage{booktabs}
\usepackage{bbding}
\usepackage{textcomp}
\usepackage{wrapfig} 
\usepackage{multirow}

\newcommand{\resultpara}[1]{\par\smallskip\noindent\textbf{#1.}\hspace{0.5em}}
\title{Gated Spatial Redundancy Projection for Pathology Transformer Attentions}

\addauthor{Zhiyuan Yang}{zhiyuan.yang@mail.concordia.ca}{1}
\addauthor{Jiahao Cheng}{jiahao.cheng@mail.concordia.ca}{1}
\addauthor{Vincent Quoc-Huy Trinh}{quoc-huy.trinh@umontreal.ca}{2}
\addauthor{Mahdi S. Hosseini\Envelope{}}{mahdi.hosseini@concordia.ca}{1,3} 

\addinstitution{
 Department of Computer Science and Software Engineering (CSSE), Concordia University, Montreal, Canada
}
\addinstitution{
 Axe Cancer, Centre de recherche du CHUM, Université de Montréal, Montréal, Canada
}
\addinstitution{
 Mila - Quebec AI Institute, Montreal, Canada
}

\runninghead{Yang et al.}{Gated Spatial Redundancy Projection}

\DeclareMathOperator{\proj}{proj}
\begin{document}

\maketitle

\begin{abstract}

Transformer models are increasingly used for whole-slide image analysis in computational pathology. Yet, WSIs differ fundamentally from natural images: neighbouring patches often contain highly similar tissue type, stain, texture, and cellular composition. We identify this \emph{local spatial redundancy} as a pathology-specific failure mode of self-attention, where dominant neighbourhood features can be repeatedly mixed into patch-tokens and weaken subtle diagnostic or prognostic deviations. We propose \textbf{Gated Spatial Redundancy Projection (Gated SRP)}, a lightweight drop-in correction module for self-attention layers. For each patch token and attention head, Gated SRP estimates a local redundancy axis from neighbouring value vectors, projects the attention output onto this axis, and applies a learned signed gate to correct the redundancy-aligned component geometrically. Across five TCGA survival cohorts, Gated SRP obtains the highest mean C-index among the compared attention variants in all cohorts, with an average improvement over the base attention, while adding only +0.02\% parameters. Across five slide-level classification datasets, it improves the base attention on 12 of 16 reported metrics and achieves the best AUC on three datasets. Code is publicly available at https://github.com/AtlasAnalyticsLab/GatedSRP.
\end{abstract}

\input{sections/introduction}
\input{sections/related_works}
\input{sections/preliminaries}
\input{sections/methodology}

\input{sections/Experiments}

\input{sections/Ablation}
\input{sections/conclusion}

\bibliography{egbib}

\appendix
\include{sections/supplementary}

\end{document}

%% file: sections/introduction.tex
\section{Introduction}
\label{sec:intro}

\begin{figure*}[ht]
  \centering
  \includegraphics[width=\textwidth]{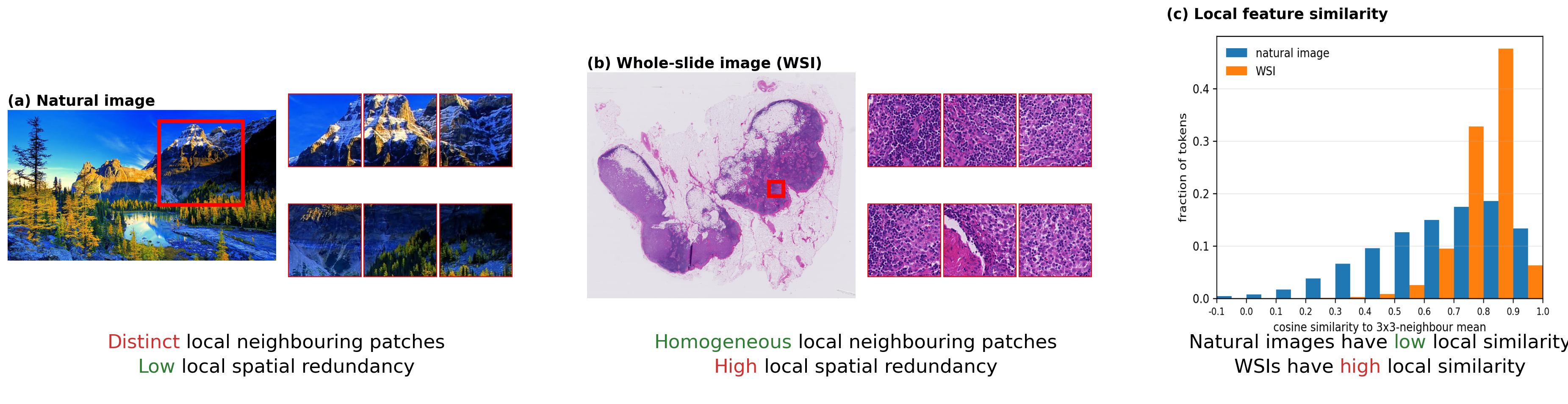}
  \caption{%
    \textbf{Local spatial redundancy in pathology.}
    (a) Nearby crops in a natural image often contain distinct semantic
    content. (b) Nearby crops in a whole-slide image (WSI) often show the same
    tissue type, stain, and texture. (c) In transformer models, after self-attention layers, each WSI patch token has a high cosine similarity to its $3\times3$-neighbour mean, whereas the corresponding similarities for natural image patches are spread far lower. We refer to this near-duplicate
    neighbourhood structure as \emph{local spatial redundancy}, which can drive transformer models to mix redundant information into diagnostically important patch tokens, and make them less distinguishable from their surrounding tissue context.%
  }
  \label{fig:local-homogeneity}
\end{figure*}

In computational pathology, transformer models are widely used to analyze giga-pixel whole-slide images (WSIs) from tissue specimens which are used for patch-level and slide-level prediction tasks~\cite{cpath}. However, unlike natural images, WSIs are often locally homogeneous within proximity of tissue regions i.e. nearby patches from the same anatomical or pathological area tend to share tissue type, stain, texture, and cellular composition~\cite{cpath, adp, adpv2}. We refer to this near-duplicate neighbourhood structure as \emph{local spatial redundancy}, illustrated in Figure~\ref{fig:local-homogeneity}. This redundancy provides useful spatial context, but it also creates a distinctive challenge in transformer models \cite{transformer, vit}. When neighbouring patch tokens carry similar visual information, the token with subtle diagnostic evidence can be easily overwhelmed by repetitive information, making it less distinguishable from other tokens after attention layers. This problem can be especially prominent in prognostic tasks such as \textbf{survival analysis}, where risk prediction often relies on weak microenvironmental cues spread across homogeneous tissue regions.

Existing research in generic transformers has shown that attention outputs, token interactions, and redundant tokens can be modified to improve model behaviour~\cite{XSA-paper,diff,integral-transformer,cog-attention,differential-gated-attention,convit,swin,neighborhood-attention,tome,evit,dynamicvit,tokenlearner,cait,specialization}. 
In computational pathology, WSI transformer models have also made strong progress in slide-level aggregation, efficient token handling, and representation learning~\cite{clam,dsmil,transmil,patch-gcn,hipt,UNI,lu2024conch,titan,gigapath,splice,tcssa,focus,pathvq}. 
However, most of these methods are not aware of the local spatial redundancy of WSIs as an image property. 
Generic transformer methods are usually designed for natural images, where redundancy is often regarded as a problem of noise, routing, efficiency, or token roles rather than as a property of the image. 
On the other hand, pathology models often use attention to aggregate slide-level evidence, but they do not explicitly address how locally repeated tissue patterns may introduce redundant information to the tokens. 
This leaves a natural question: \textit{can we make transformer attention more suitable for pathology by making patch-token updates aware of local spatial redundancy?}

We propose \textbf{Gated Spatial Redundancy Projection (Gated SRP)}, a
lightweight drop-in module for attention layers. For each patch token and attention
head, Gated SRP estimates a local spatial redundancy direction from
neighbouring value vectors, projects the attention output onto this direction,
and uses a small signed gate to decide how much of the projected component
should be corrected. The module is applied
only to patch tokens, leaving the \textsc{cls} token unchanged for slide-level
aggregation. It can be inserted after any attention layers without changing the
attention routing, classifier, or loss function. Furthermore, Gated SRP is designed as a safe low-overhead correction rather than a large
new prediction module. On average, it adds only a lightweight gate with \textasciitilde{0.02\%} of total parameters. The base attention
remains exactly reachable throughout training: when the learned correction
coefficient is zero, the layer reduces to the original attention layer. This
identity path makes the module safe to integrate after attention blocks,
because the model can suppress the spatial correction whenever the baseline
attention update is preferable. Our contributions are as follows:



\begin{itemize}\itemsep2pt\leftmargin=1.4em

  \item \textbf{A novel self-attention correction method for pathology.}
  We introduce Gated Spatial Redundancy Projection (Gated SRP) to correct the patch tokens, following self-attention layers, to mitigate the risk of repetitive patterns within the proximity of the tissue region in WSI, which can overwhelm subtle diagnostic evidence in each token.

  \item \textbf{A pathology-driven view of transformer attention.}
  We identify local spatial redundancy as a potential failure mode of self-attention in pathology transformers, helping to explain why attention in highly homogeneous WSIs can differ from natural images.

  \item \textbf{Flexibility, efficiency, and stability.}
  Gated SRP is a lightweight drop-in module that can be inserted after any self-attention block with minimal additional parameters. The module is designed to be able to recover the base attention layer when necessary, which makes the performance on par with the base attention in the worst-case scenario.

  \item \textbf{A systematic evaluation of the Gated SRP mechanism.}
  We evaluate the Gated SRP on five TCGA survival cohorts and five slide-level classification datasets, where Gated SRP achieves favorable performance across survival and classification settings, with ablations supporting the contribution of the proposed spatial correction mechanism.

\end{itemize}

%% file: sections/related_works.tex
\section{Related Works}
\label{sec:related-work}

\resultpara{Attention correction and token roles}Recent work shows that attention can contain redundant or noisy components that should be explicitly corrected. Exclusive Self-Attention (XSA) identifies token self-information as redundant information and explicitly removes self-information from each token because it occupies the token's capacity to carry contextual information~\cite{XSA-paper}. Differential Transformer subtracts paired attention maps to reduce common attention noise~\cite{diff}, and related methods further study negative, integral, or gated forms of attention correction~\cite{cog-attention,integral-transformer,differential-gated-attention}. These works suggest attention is not always best used as-is. However, these methods correct attention outputs in a generic way and are not built from the spatial neighbourhoods of pathology slides. Another related direction studies the different roles of patch tokens and the \textsc{CLS} token. CaiT delays \textsc{CLS}-token interaction, and recent token-specialization work shows that patch tokens and readout tokens may benefit from different processing rules~\cite{cait,specialization}. Similarly, pathology transformer models must preserve local patch evidence while still producing a global slide representation, suggesting attention corrections should be done differently on patch tokens and the \textsc{CLS} token.

\resultpara{Locality and token redundancy in vision transformers}Another line of work also studies locality and redundant tokens in Vision Transformers, which relates closely to local spatial redundancy in WSIs. ConViT introduces a soft convolutional bias into attention, while Swin Transformer and Neighborhood Attention Transformer restrict attention to local windows or neighbourhoods~\cite{convit,swin,neighborhood-attention}. These methods show that local image structure is useful in natural images. However, this contradicts the local spatial redundancy property of WSIs. These methods encourage/limit the tokens to attend to neighbour tokens, which leads to redundant token information being explicitly mixed together and potentially harms the downstream performance in pathology tasks. Other methods treat redundancy as an efficiency problem: DynamicViT and EViT remove less useful tokens, TokenLearner summarizes an image with a small set of learned tokens, and ToMe merges similar tokens to reduce computation~\cite{dynamicvit,evit,tokenlearner,tome}. These methods recognize that many visual tokens can be repeated or less informative. However, they usually prune, merge, summarize, or reroute tokens, which sometimes destabilize model training through changed token behavior and loss of information. 


\resultpara{Whole-slide image modelling in computational pathology}In computational pathology, many WSI models focus on learning slide-level representations from large sets of patches. MIL methods such as CLAM and DSMIL identify important regions for slide-level prediction~\cite{clam,dsmil}. Transformer and graph-based methods such as TransMIL, Patch-GCN, and HIPT further model spatial or hierarchical relations among tissue regions~\cite{transmil,patch-gcn,hipt}. Pathology foundation models such as UNI, CONCH, TITAN, GigaPath and MOOZY provide strong patch-, slide-, or patient-level representations that can be transferred to downstream tasks~\cite{UNI,lu2024conch,titan,gigapath,moozy}. More recent methods address WSI scale by selecting representative patches, filtering tokens, or compressing slide representations~\cite{splice,focus,tcssa,pathvq}. These works have greatly improved WSI analysis, but they mostly use attention for aggregation and representation learning, or efficiency. They do not directly study how local common tissue patterns can affect attention updates and weaken subtle diagnostic or prognostic signals.


%% file: sections/preliminaries.tex
\section{Methodology}
\subsection{Preliminaries}
\label{sec:prelim}

In standard vision transformers \cite{vit}, a transformer block takes a set of $N$ visual tokens $\{x_i\}_{i=1}^{N}$, $x_i\in \mathbb{R}^{D}$ as inputs, optionally preceded by a learned classification token $x_{\text{\textsc{cls}}}$. Each block uses multi-head self-attention with $H$ heads of head dimension $d=D/H$. For a single head we drop the head index in this subsection; per-head matrices are introduced explicitly only when the distinction matters.

\resultpara{Standard self-attention (SA)}For tokens $\{x_i\}$, projections $q_i = W_q x_i,\;\; k_j = W_k x_j,\;\; v_j = W_v x_j \in \mathbb{R}^{d}$, attention weights and output are
\begin{equation}
\label{eq:sa-weights}
a_{i,j} \;=\; \frac{\exp\!\left(q_i^{\top}k_j / \sqrt{d}\right)}
                   {\sum_{j'} \exp\!\left(q_i^{\top}k_{j'} / \sqrt{d}\right)},
\qquad
y_i \;=\; \sum_{j} a_{i,j}\, v_j .
\end{equation}
A residual connection then adds $x_i$ back at the block boundary, so the
token's own information is preserved at the block output regardless of
what the attention contribution $y_i$ adds. The SA contribution $y_i$
is the object the post-attention modifications below operate on.

\resultpara{Exclusive Self-Attention (XSA)}\citet{XSA-paper} observed that under standard SA, the output $y_i$ frequently retains a non-trivial component along the token's \emph{own} value vector $v_i$, even when the diagonal weight $a_{i,i}$ is small. This happens either through residual paths or via correlated $\{v_j\}$ across the softmax weights $a_{i,\cdot}$. Over-emphasizing token self-information limits the model's ability to gather \emph{contextual} information from other tokens, thus harming the model's performance. To force the attention update to carry strictly contextual information, XSA replaces $y_i$ with its projection onto the subspace orthogonal to $v_i$:
\begin{equation}
\label{eq:xsa}
\boxed{\;\;
z_i \;=\; y_i \;-\; \frac{y_i^{\top} v_i}{\|v_i\|_2^{2}}\, v_i
 = y_i - \proj_{v_i}y_i\;\;}
\qquad
\text{(per head; \citealp{XSA-paper}, Eq.~2)}
\end{equation}
Equivalently, with $\hat v_i = v_i / \|v_i\|_2$,
\begin{equation}
\label{eq:xsa-unit}
z_i \;=\; y_i \;-\; (y_i^{\top}\hat v_i)\,\hat v_i,
\end{equation}
i.e.\ exact removal of the component of $y_i$ along the unit value
vector $\hat v_i$. This orthogonal projection separates \emph{where
attention is allocated} from \emph{which direction of the aggregated
value is retained}. The model may still attend to token $i$, but the
$\hat v_i$-aligned component is removed after aggregation. It is often preferred over token removal or attention masking because the latter changes the routing problem itself
and can discard useful orthogonal context. Therefore, projection
preserves the routing capacity of the attention while leaving self-like,
point-wise content to the residual/FFN path, matching the division of
labour motivating XSA.

%% file: sections/methodology.tex
\subsection{Motivation}
\label{sec:motivation}

XSA provides a simple and intuitive approach to remove certain information from attention output tokens. However, WSIs and their image patches have three key properties that constrain token-modification mechanisms: %
\begin{itemize}\itemsep2pt\leftmargin=1.4em
  \item \textbf{Redundancy is local spatial information.} Adjacent patch tokens inside tumour,
  stroma, or lymphoid regions often share cell type, staining, and
  texture, so the relevant redundancy axis is local neighbourhood
  context rather than the token's own value alone.

  \item \textbf{Projection strength should be token-dependent.} A patch token inside
  homogeneous tissue may tolerate strong projection, whereas an isolated
  tumour focus may contain the slide's entire diagnostic signal.

  \item \textbf{Rare evidence may require amplification.} For isolated
  or boundary tokens, the useful signal is their disagreement with the
  neighbourhood; one-sided removal can suppress exactly that evidence.
\end{itemize}
XSA's three coupled choices (projection along $\hat v_i$, a fixed coefficient uniformly applied to all tokens, and one-sided removal) directly contradict the three properties above. In the next subsection, we propose the \textbf{gated spatial redundancy projection} attention, which addresses the three properties directly.

\subsection{Gated Spatial Redundancy Projection (Gated SRP)}
\label{sec:gated-srp}
\label{sec:signed-gate}
\begin{figure*}[ht]
  \centering
  \includegraphics[width=0.8\textwidth]{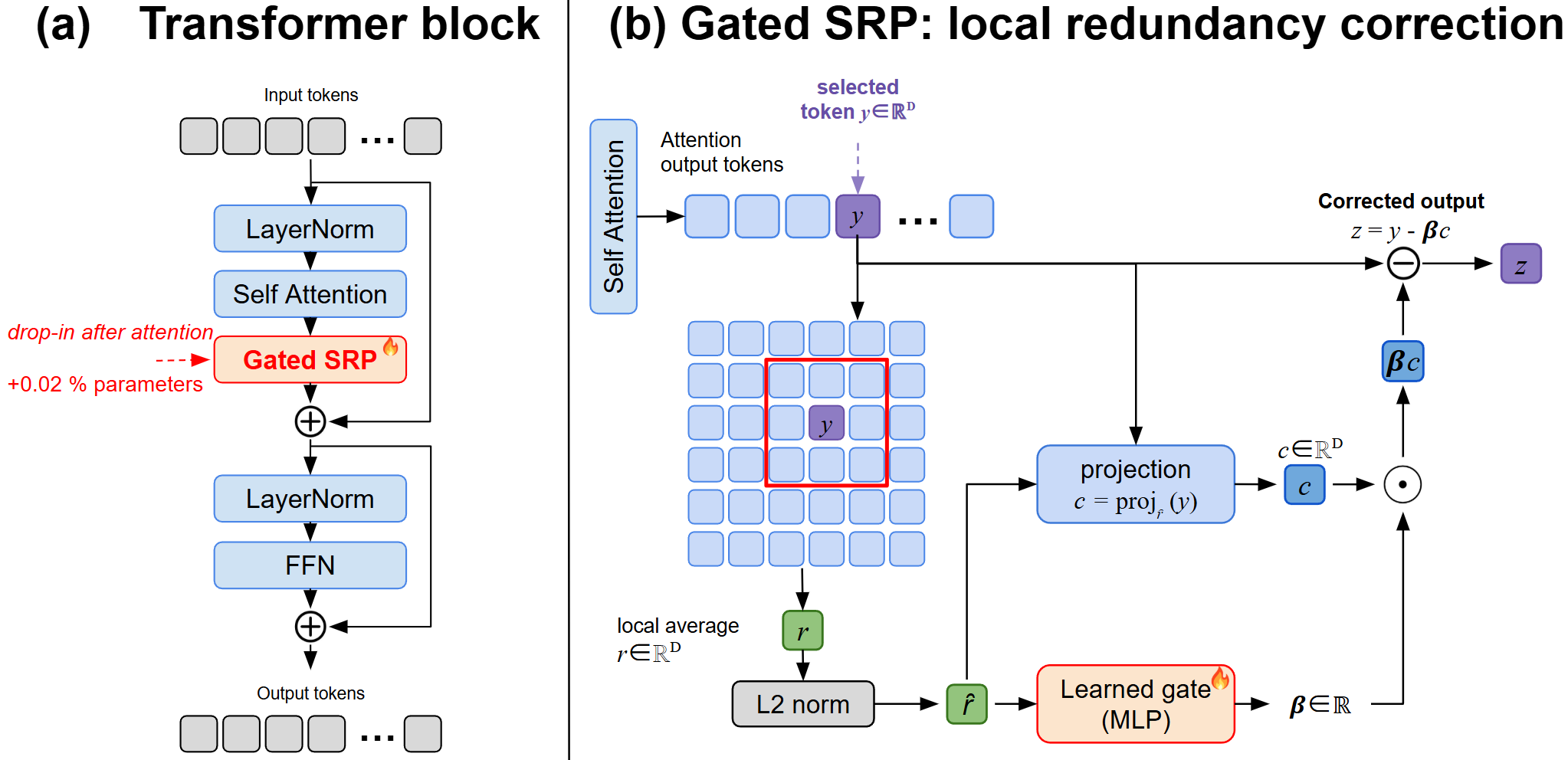}
  \caption{
  \textbf{Method overview.} (a) Gated SRP is a lightweight drop-in module inserted after a self-attention layer, adding only \textasciitilde{}0.02\% parameters.
(b) For each patch-token attention output $y$, Gated SRP uses the local $3 \times 3$ neighbourhood to estimate a redundancy direction $\hat r$.
It projects $y$ onto this direction to obtain the neighbourhood-aligned component $c=\mathrm{proj}_{\hat r}(y)$.
A learned gate predicts a signed coefficient $\beta$, and the corrected output is $z=y-\beta c$.
The correction is applied to all patch tokens but not to the \textsc{cls} token.
  }
  \label{fig:overview}
\end{figure*}

As illustrated in Figure~\ref{fig:overview}, Gated SRP is a drop-in attention correction module after any self-attention layers. It modifies only the
attention contribution $y_{i,h}$ of a visual token $i$ in head $h$.
It first defines the local common component
\begin{equation}
\label{eq:common-component}
c_{i,h} = \proj_{r_{i,h}}y_{i,h} \;=\; \bigl(y_{i,h}^{\top}\hat r_{i,h}\bigr)\hat r_{i,h},
\end{equation}
where $r_{i,h}$ is the \textbf{local spatial redundancy vector} and $\hat r_{i,h}$ is the L2-normalized $r_{i,h}$ that represents the \textbf{spatial-neighbourhood redundancy axis}, as shown in Eq \ref{eq:r-def}. Then, the local common component is multiplied by a signed, token-specific coefficient $\beta_{\mathrm{eff},\,i,h}$ (Eq~\ref{eq:beta-eff}), and subtracted from $y_{i,h}$ to correct it:
\begin{equation}
\label{eq:gated-srp}
\boxed{\;\;
z_{i,h} \;=\; y_{i,h} \;-\; \beta_{\mathrm{eff},\,i,h}\,c_{i,h}
\;\;}
\qquad\text{(per head)}
\end{equation}
This single equation implements the three changes motivated above. First,
the axis $\hat r$ is spatial rather than self-referential. Second, the strength
$\beta_{\mathrm{eff}}$ varies per token and head. Lastly, the signed range
of $\beta_{\mathrm{eff}}$ lets the model choose anti-projection,
identity, projection, or reflection as local context requires. We explain each component and design choices in later paragraphs, with a PyTorch-style pseudocode block provided in the appendix.

\resultpara{Local spatial redundancy axis}We define local spatial redundancy as the local common pattern within a spatial neighbourhood. For each visual token $i$, let $\mathcal{N}(i)$ be the set of valid spatial neighbours on the patch grid within a $n\times n$ window, where $n$ is a hyperparameter. We use $n=3$, which gives an 8-neighbour grid when coordinates are available, and the analogous fixed grid for patch-level images. The head-$h$ neighbourhood mean is
\begin{equation}
\label{eq:r-def}
r_{i,h} \;=\; \frac{1}{m_i}\sum_{j\in\mathcal{N}(i)}
              \mathrm{StopGrad}(v_{j,h}),
\qquad
\hat r_{i,h} \;=\; r_{i,h}\,\big/\,\bigl(\|r_{i,h}\|_2 + \epsilon\bigr),
\qquad
m_i = |\mathcal{N}(i)|,
\end{equation}
where $\mathrm{StopGrad}(\cdot)$ denotes the stop-gradient operator and
$\epsilon$ is a small stability constant. We stop gradients through
$\hat r_{i,h}$ so that the neighbour mean is a geometric reference
rather than a trainable escape route through $W_v$. Therefore, the learnable part of the method is the decision of how to use this local axis,
not the axis itself.

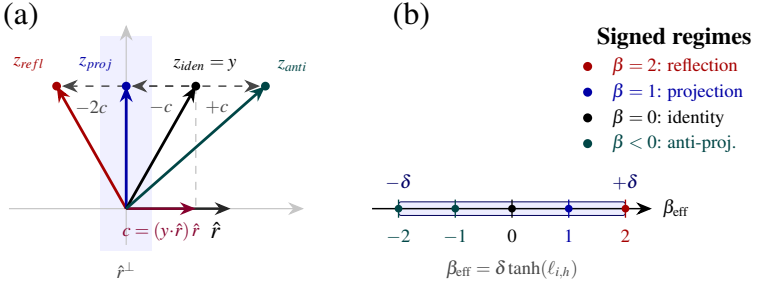
\begin{figure}[ht]
\centering
\begin{tikzpicture}[
  >={Stealth[length=2.1mm,width=1.7mm]},
  font=\small,
  axis/.style={->, line width=0.45pt, color=gray!45},
  ref/.style={->, line width=0.95pt, color=black!85},
  regarr/.style={->, line width=1.05pt},
  guide/.style={dashed, line width=0.45pt, color=gray!50},
  modarr/.style={->, dashed, line width=0.55pt, color=gray!55!black,
                 shorten <=2.5pt, shorten >=2pt},
  legenddot/.style={circle, fill, inner sep=1.45pt},
]

\begin{scope}
  \node[font=\large, anchor=west] at (-1.75,2.55) {(a)};

  \fill[blue!6] (-0.34,-0.55) rectangle (0.34,2.28);
  \draw[axis] (-1.55,0) -- (2.35,0);
  \draw[axis] (0,-0.48) -- (0,2.43);
  \node[font=\scriptsize, color=gray!55!black, anchor=north]
    at (0,-0.58) {$\hat r^{\perp}$};

  \draw[ref] (0,0) -- (1.38,0);
  \node[font=\small, anchor=north] at (1.18,-0.05) {$\hat r$};

  \draw[->, line width=0.85pt, color=purple!70!black] (0,0) -- (0.92,0);
  \node[font=\scriptsize, color=purple!70!black, anchor=north]
    at (0.46,-0.05) {$c=(y\!\cdot\!\hat r)\,\hat r$};

  \draw[guide] (0.92,1.62) -- (0.92,0);

  \draw[regarr, color=red!65!black]       (0,0) -- (-0.92,1.62);
  \draw[regarr, color=blue!65!black]      (0,0) -- (0,1.62);
  \draw[regarr, color=black]              (0,0) -- (0.92,1.62);
  \draw[regarr, color=teal!55!black]      (0,0) -- (1.84,1.62);

  \draw[modarr] (0.92,1.62) -- (0.00,1.62);   
  \draw[modarr] (0.92,1.62) -- (-0.92,1.62);  
  \draw[modarr] (0.92,1.62) -- (1.84,1.62);   

  \fill[red!65!black]  (-0.92,1.62) circle (1.65pt);
  \fill[blue!65!black] (0,1.62) circle (1.65pt);
  \fill[black]         (0.92,1.62) circle (1.65pt);
  \fill[teal!55!black] (1.84,1.62) circle (1.65pt);

  \node[font=\scriptsize, color=red!65!black, anchor=south east]
    at (-0.95,1.70) {$z_{refl}$};
  \node[font=\scriptsize, color=blue!65!black, anchor=south east]
    at (-0.05,1.70) {$z_{proj}$};
  \node[font=\scriptsize, color=black, anchor=south west]
    at (0.5,1.70) {$z_{iden}=y$};
  \node[font=\scriptsize, color=teal!55!black, anchor=south west]
    at (1.88,1.70) {$z_{anti}$};

  \node[font=\scriptsize, color=gray!55!black, anchor=north]
    at (0.46,1.55) {$-c$};
  \node[font=\scriptsize, color=gray!55!black, anchor=north]
    at (-0.46,1.55) {$-2c$};
  \node[font=\scriptsize, color=gray!55!black, anchor=north]
    at (1.20,1.55) {$+c$};
\end{scope}

\begin{scope}[xshift=3.75cm, yshift=0cm]
  \node[font=\large, anchor=west] at (-0.35,2.55) {(b)};

  \node[font=\small\bfseries, anchor=west] at (2.35,2.25) {Signed regimes};

  \fill[red!65!black] (2.35,1.88) circle (1.7pt);
  \node[font=\scriptsize, color=red!65!black, anchor=west]
    at (2.55,1.88) {$\beta=2$: reflection};

  \fill[blue!65!black] (2.35,1.54) circle (1.7pt);
  \node[font=\scriptsize, color=blue!65!black, anchor=west]
    at (2.55,1.54) {$\beta=1$: projection};

  \fill[black] (2.35,1.20) circle (1.7pt);
  \node[font=\scriptsize, color=black, anchor=west]
    at (2.55,1.20) {$\beta=0$: identity};

  \fill[teal!55!black] (2.35,0.86) circle (1.7pt);
  \node[font=\scriptsize, color=teal!55!black, anchor=west]
    at (2.55,0.86) {$\beta<0$: anti-proj.};

  \draw[fill=blue!7, draw=blue!35!black, line width=0.35pt,
        rounded corners=1pt] (-0.15,-0.09) rectangle (2.85,0.09);
  \draw[->, line width=0.55pt] (-0.50,0) -- (3.20,0);
  \node[font=\scriptsize, anchor=west] at (3.22,0) {$\beta_{\mathrm{eff}}$};

  \foreach \x/\col/\lab in {%
      -0.15/teal!55!black/$-2$,
       0.60/teal!55!black/$-1$,
       1.35/black/$0$,
       2.10/blue!65!black/$1$,
       2.85/red!65!black/$2$%
  }{%
    \draw[\col, line width=0.5pt] (\x,0.12) -- (\x,-0.12);
    \fill[\col] (\x,0) circle (1.45pt);
    \node[font=\scriptsize, color=\col, anchor=north] at (\x,-0.15) {\lab};
  }

  \node[font=\scriptsize, color=blue!40!black, anchor=south]
    at (-0.15,0.13) {$-\delta$};
  \node[font=\scriptsize, color=blue!40!black, anchor=south]
    at (2.85,0.13) {$+\delta$};

  \node[font=\scriptsize, color=gray!55!black, anchor=north]
    at (1.35,-0.55)
    {$\beta_{\mathrm{eff}}=\delta\tanh(\ell_{i,h})$};
\end{scope}

\end{tikzpicture}

\caption{\textbf{Geometric illustration of the signed gate.}
\textbf{(a)} For a fixed local redundancy axis $\hat r$, the redundant
component $c=(y\cdot\hat r)\,\hat r$ lies along $\hat r$, and
the signed gate writes $z = y - \beta_{\mathrm{eff}}\,c$. Varying
$\beta_{\mathrm{eff}}$ therefore moves the output only along $\hat r$. The identity endpoint is labelled
$y=z_{iden}$. The special cases are labelled as $z$ variants:
projection, reflection, and anti-projection (amplification).
\textbf{(b)} The signed gate places these regimes on the bounded range
$\beta_{\mathrm{eff}}=\delta\tanh(\ell_{i,h})$, with the regime selected
separately for each token and head.}
\label{fig:gated-srp-regimes}
\end{figure}

\resultpara{Signed local gate}The scalar $\beta_{\mathrm{eff},\,i,h}$ is produced from a gate logit $\ell_{i,h}$ (Eq~\ref{eq:gate-logit}) by a bounded signed head:
\begin{equation}
\label{eq:beta-eff}
\beta_{\mathrm{eff},\,i,h} \;=\; \delta\cdot\tanh(\ell_{i,h})
\;\in\; (-\delta,\,+\delta).
\end{equation}
where $\delta$ is a hyperparameter controlling the range of $\beta_{\mathrm{eff},\,i,h}$. Although $\beta_{\mathrm{eff}}$ is a real number that allows partial geometric corrections, certain $\beta_{\mathrm{eff}}$ values have very specific geometric meanings, as shown in Fig \ref{fig:gated-srp-regimes} and the table below:
\[
\begin{array}{c|c}
\beta_{\mathrm{eff}} & \text{operation on }c_{i,h} \\ \hline
0 & \text{identity: } z=y \\
1 & \text{projection: remove the } \hat r\text{-component} \\
2 & \text{reflection: flip the } \hat r\text{-component} \\
<0 & \text{anti-projection: amplify alignment with }\hat r .
\end{array}
\]
We use $\delta\in\{0.5,1,1.5, 2\}$ as a small architectural choice: $\delta=1$ allows identity, projection, and anti-projection, $\delta=2$ also allows reflection, while $0.5$ and $1.5$ are kept as the middle ground in-between. Keeping $\delta$ fixed makes this representational choice explicit rather than hiding it inside optimization.

\resultpara{Factored gate logit}The logit combines token-level context with head-level alignment:
\begin{equation}
\label{eq:gate-logit}
\ell_{i,h} \;=\;
  \underbrace{g_{\text{tok}}(d^{\text{tok}}_i)}_{\text{shared across heads}}
  \;+\;
  \underbrace{w_h^{\top}d^{\text{head}}_{i,h} + b_h}_{\text{head-specific}}
  \;+\;
  \underbrace{b_{l,h}}_{\text{per-(layer, head) bias}}\,,
\end{equation}
where $g_{\text{tok}}$ is a two-layer MLP with GELU activation, and $d^{\text{tok}}_i\in\mathbb{R}^{T_t}$ and $d^{\text{head}}_{i,h}\in\mathbb{R}^{T_h}$ are the token/head-level diagnostics vectors. The diagnostic vectors are defined as
\newline
\begin{equation}
\label{eq:diag-tok}
d^{\text{tok}}_i \;=\; \bigl[\,
  h_{\text{local},\,i},\;\;
  m_i/8,\;\;
  \log(1+m_i)
\,\bigr],
\end{equation}
\begin{equation}
\label{eq:diag-head}
d^{\text{head}}_{i,h} \;=\; \bigl[\,
  \cos(y_{i,h},\hat r_{i,h}),\;\;
  |\cos(y_{i,h},\hat r_{i,h})|,\;\;
  \log(1+\|y_{i,h}\|_2)
\,\bigr],
\end{equation}
with $T_t=T_h=3$ in the default gate. The token features summarise
whether the local tissue context is reliable enough to project
against:
\begin{itemize}\itemsep1pt\leftmargin=1.4em
  \item $h_{\text{local},\,i}=m_i^{-1}\sum_{j\in\mathcal{N}(i)}
        \cos(\phi_i,\phi_j)$ is the mean cosine similarity between the
        raw patch embedding $\phi_i$ and its neighbours, so it
        measures how homogeneous the surrounding tissue actually is.
  \item $m_i/8$ is the linear fraction of valid neighbours, marking
        boundary and corner tokens whose spatial context is
        incomplete.
  \item $\log(1+m_i)$ is a log-confidence proxy for $\hat r_{i,h}$:
        the neighbour mean variance scales as $1/m_i$, so this term lets the gate shrink $\beta_{\rm eff}$ when the axis is built from few samples.
\end{itemize}
The head features describe how the current attention contribution
$y_{i,h}$ relates to that axis:
\begin{itemize}\itemsep1pt\leftmargin=1.4em
  \item $\cos(y_{i,h},\hat r_{i,h})$ is the signed alignment of the
        head output with the redundancy axis, indicating whether
        subtracting $c_{i,h}$ acts as projection or anti-projection.
  \item $|\cos(y_{i,h},\hat r_{i,h})|$ is the unsigned alignment
        magnitude, separating how much of $y_{i,h}$ lies along
        $\hat r_{i,h}$ from which direction it points.
  \item $\log(1+\|y_{i,h}\|_2)$ is the log-magnitude of the
        contribution being projected, so the gate can avoid
        corrections on tokens whose attention output is essentially
        zero.
\end{itemize}

The factorization of the gate logit is intentional: the shared token term decides whether the local tissue context is redundant; the per-head term decides how strongly this head's attention output aligns with that context, and $b_{l,h}$ absorbs layer/head-specific scale shifts. This keeps the gate interpretable and small, with $\mathcal{O}(T_t h_{\text{mlp}} + H T_h)$ parameters per active block. In our settings, this contributes less than $0.02\%$ of the
trainable parameter budget. The method is therefore primarily an
inductive-bias change rather than a substantial capacity increase.

\resultpara{Identity initialization}The output layer of $g_{\text{tok}}$, all per-head weights
$\{w_h,b_h\}$, and all layer/head biases $b_{l,h}$ are initialized to
zero. The hidden layer of $g_{\text{tok}}$ keeps its standard
initialization. Consequently $\ell_{i,h}=0$ and
$\beta_{\mathrm{eff},\,i,h}=0$ for every token and head at step~1, so
Eq.~\eqref{eq:gated-srp} exactly recovers standard self-attention.
Any non-zero projection is therefore learned during optimization rather
than injected by the architecture at initialization.

\resultpara{Default gradient convention}Both diagnostic vectors $d^{\text{tok}}_i$ and $d^{\text{head}}_{i,h}$ are detached before entering the gate. Thus, gate parameters are trained through their effect on $z_{i,h}$, but the upstream attention projection does not receive an additional gradient path through feature construction. This separates the specified mechanism from the live-input ablation, where gradients through the diagnostic features are allowed and analyzed separately.

\resultpara{Application scope}
Gated SRP is applied only to visual patch tokens. The
\textsc{cls} token is left unchanged due to the lack of spatial neighbours. When the classifier reads only
\textsc{cls}, patch-token projection in the final attention block has
no same-block path to the loss, so we omit the gate in that final
block. The modified heads are concatenated and passed through the
standard output projection and residual connection exactly as in the
baseline transformer.

%% file: sections/Experiments.tex
\section{Experiments}
\label{sec:experiments}

\subsection{Experiment Settings}
\label{sec:exp-settings}
We evaluate Gated SRP on four public WSI classification datasets, CAMELYON16/17\cite{cam16,cam17}, BRACS\cite{bracs}, and PANDA\cite{panda}, one in-house ADPv2-derived KGH WSI subset~\cite{adpv2}, and five TCGA survival cohorts, TCGA-KIRC, KIRP, LUAD, STAD, and UCEC~\cite{NCI_TCGA}. We extract non-overlapping tissue patches of size $256\times256$ pixels from all WSIs using AtlasPatch \cite{atlaspatch} at $20\times$ magnification. All patches are pre-embedded into feature vectors using UNI-v2 \cite{UNI}. Since WSIs often lead to exploding memory consumption due to their large sizes, we use a four-layer TransMIL-style Nystr\"om transformer aggregator as the base architecture instead of a dense ViT \cite{vit,transmil,nystrom}. To recreate complete transformer blocks as in modern ViT models, we add the missing pre-norm FFN sub-block after the TransMIL transformer layers. For evaluation and comparisons, we integrate or replace the base attention layers with different methods. Specifically, we compare our method against the base Nystr\"om Attention (NA) \cite{nystrom}, XSA \cite{XSA-paper}, and differential attention (Diff) \cite{diff}:
\begin{enumerate}
    \item \textbf{Nystr\"om Attention (NA)} is the base attention used in the TransMIL architecture to solve the exploding memory issue in WSIs \cite{transmil, nystrom}. It approximates the standard MHSA using landmark tokens and low-rank kernel approximation~\cite{vit, transformer}.
    \item \textbf{XSA} is a drop-in module after attention layers that removes the component of each attention output aligned with the token's own value direction, enforcing a more explicitly contextual attention update~\cite{XSA-paper}.
    \item \textbf{Diff} modifies a base attention layer by using paired attention modules, which suppress common-mode attention patterns by subtracting the paired attention maps before forming the attention output~\cite{diff}. The subtraction can lead to negative attentions being allocated to certain tokens, which actively removes their information from others.
\end{enumerate}
All experiments are conducted under the same five random seeds that control the dataset split, model initialization, and training. All compared methods share the same training and evaluation protocol under paired internal dataset splits. More dataset and implementation details are provided in the appendix.

\subsection{Quantitative Results}

\resultpara{Survival Analysis}
We first evaluate Gated SRP on five TCGA survival cohorts~\cite{NCI_TCGA}: KIRC, KIRP, LUAD, STAD, and UCEC, using the concordance index (C-index) to measure whether predicted risk scores correctly rank patients by survival outcome. Table~\ref{tab:survival-main} shows
\begin{wraptable}{r}{0.56\linewidth}
\centering
\caption{\textbf{Survival analysis results.}
Mean case-level C-index over five random seeds. Green/red indicates increase/decrease relative to the base NA.}
\label{tab:survival-main}
{\scriptsize
\setlength{\tabcolsep}{2.4pt}
\renewcommand{\arraystretch}{1.04}
\resizebox{\linewidth}{!}{%
\begin{tabular}{@{}lccccc@{}}
\toprule
\textbf{Method} & \textbf{KIRC} & \textbf{KIRP} & \textbf{LUAD} & \textbf{STAD} & \textbf{UCEC} \\
\midrule
NA~\cite{transmil, nystrom}
& 0.7110 & 0.7247 & 0.5513 & 0.5910 & 0.6756 \\
XSA~\cite{XSA-paper}
& \textcolor[HTML]{137333}{0.7192}
& \textcolor[HTML]{B3261E}{0.6795}
& \textcolor[HTML]{B3261E}{0.5359}
& \textcolor[HTML]{137333}{0.6035}
& \textcolor[HTML]{B3261E}{0.6233} \\
Diff~\cite{diff}
& \textcolor[HTML]{137333}{0.7241}
& \textcolor[HTML]{B3261E}{0.7020}
& \textcolor[HTML]{137333}{0.5546}
& \textcolor[HTML]{B3261E}{0.5847}
& \textcolor[HTML]{B3261E}{0.6489} \\
\textbf{Gated SRP}
& \textcolor[HTML]{137333}{\textbf{0.7257}}
& \textcolor[HTML]{137333}{\textbf{0.7648}}
& \textcolor[HTML]{137333}{\textbf{0.5832}}
& \textcolor[HTML]{137333}{\textbf{0.6171}}
& \textcolor[HTML]{137333}{\textbf{0.6973}} \\
\bottomrule
\end{tabular}}}
\end{wraptable}
that Gated SRP obtains the highest mean C-index among the compared attention variants on all five cohorts. Compared with the base NA baseline, the mean C-index improves across all cohorts, although the magnitude of the improvement varies across datasets. The mixed behaviour of XSA is consistent with our motivation that redundancy in WSIs may be better captured by local spatial neighbourhoods than by a token's own value direction alone. Removing the component aligned with a token's own value direction is not as reliable as correcting the component aligned with its local tissue neighbourhood. The gains from our method are also consistent with the biological nature of survival prediction, where prognostic risk often depends on weak and spatially distributed cues from the tumour microenvironment. Our method leverages the information from local patch neighbourhoods, which provides a coarse spatial proxy for these microenvironmental relationships. This suggests that correcting the neighbourhood-aligned component may help reduce homogeneous local context while retaining deviations that are useful for prognostic modelling.



\resultpara{WSI Classification}We further evaluate our Gated SRP on five slide-level classification datasets. We use F1, accuracy (Acc), and AUC for classification performance, with quadratic-weighted Cohen's kappa $\kappa_q$ for PANDA. Table~\ref{tab:main-classification} shows that Gated SRP gives the most favourable overall mean performance relative to the base NA across the reported metrics, while XSA and Diff show more dataset-dependent behaviour. As a lightweight drop-in module, Gated SRP improves the base NA baseline on $12$ of the $16$ reported classification metrics, achieves the best mean AUC on three of the five datasets, and obtains the best F1 and accuracy on KGH and BRACS. On CAMELYON17, our method is close to the base NA overall, whereas the other attention variants achieve lower mean performance.

\begin{table*}[ht]
\centering
\caption{\textbf{Classification results.} We report F1, accuracy, and AUC for all datasets. PANDA additionally reports quadratic-weighted Cohen's kappa. Green and red indicate meaningful increase and decrease relative to the base NA. The highest score for each column is \textbf{bolded}.}
\label{tab:main-classification}
{\scriptsize
\setlength{\tabcolsep}{2.1pt}
\renewcommand{\arraystretch}{1.04}
\resizebox{\textwidth}{!}{%
\begin{tabular}{l*{16}{c}}
\toprule
\textbf{Method} & \multicolumn{3}{c}{\textbf{CAM16}} & \multicolumn{3}{c}{\textbf{CAM17}} & \multicolumn{3}{c}{\textbf{KGH}} & \multicolumn{4}{c}{\textbf{PANDA}} & \multicolumn{3}{c}{\textbf{BRACS}} \\
\cmidrule(lr){2-4}\cmidrule(lr){5-7}\cmidrule(lr){8-10}\cmidrule(lr){11-14}\cmidrule(lr){15-17}
& \textit{F1} & \textit{Acc} & \textit{AUC} & \textit{F1} & \textit{Acc} & \textit{AUC} & \textit{F1} & \textit{Acc} & \textit{AUC} & $\mathit{\kappa_q}$ & \textit{F1} & \textit{Acc} & \textit{AUC} & \textit{F1} & \textit{Acc} & \textit{AUC} \\
\midrule
NA~\cite{nystrom} & 0.9731 & 0.9741 & 0.9974 & \textbf{0.7836} & \textbf{0.9077} & 0.9470 & 0.8297 & 0.8337 & 0.9665 & 0.8822 & 0.6861 & 0.7374 & 0.9313 & 0.4179 & 0.5439 & \textbf{0.8472} \\
XSA~\cite{XSA-paper} & 0.9730 & 0.9741 & 0.9977 & \textcolor[HTML]{B3261E}{0.7342} & \textcolor[HTML]{B3261E}{0.8916} & 0.9441 & \textcolor[HTML]{137333}{0.8469} & \textcolor[HTML]{137333}{0.8482} & 0.9642 & 0.8811 & 0.6854 & \textcolor[HTML]{137333}{0.7397} & 0.9289 & \textcolor[HTML]{137333}{0.4328} & \textcolor[HTML]{137333}{0.5567} & 0.8425 \\
Diff~\cite{diff} & \textcolor[HTML]{137333}{\textbf{0.9923}} & \textcolor[HTML]{137333}{\textbf{0.9926}} & 0.9972 & \textcolor[HTML]{B3261E}{0.7578} & \textcolor[HTML]{B3261E}{0.8976} & \textcolor[HTML]{B3261E}{0.9303} & \textcolor[HTML]{137333}{0.8471} & \textcolor[HTML]{137333}{0.8482} & 0.9688 & 0.8826 & 0.6787 & 0.7322 & \textcolor[HTML]{137333}{\textbf{0.9345}} & \textcolor[HTML]{B3261E}{0.3874} & 0.5366 & \textcolor[HTML]{B3261E}{0.8325} \\
\midrule
\textbf{Gated SRP} & \textcolor[HTML]{137333}{0.9808} & \textcolor[HTML]{137333}{0.9815} & \textcolor[HTML]{137333}{\textbf{0.9986}} & 0.7785 & 0.9015 & \textcolor[HTML]{137333}{\textbf{0.9504}} & \textcolor[HTML]{137333}{\textbf{0.8544}} & \textcolor[HTML]{137333}{\textbf{0.8566}} & \textcolor[HTML]{137333}{\textbf{0.9695}} & \textbf{0.8842} & \textcolor[HTML]{137333}{\textbf{0.6932}} & \textcolor[HTML]{137333}{\textbf{0.7461}} & 0.9292 & \textcolor[HTML]{137333}{\textbf{0.4424}} & \textcolor[HTML]{137333}{\textbf{0.5603}} & 0.8454 \\
\bottomrule
\end{tabular}}}
\end{table*}

\begin{figure*}[ht]
  \centering
  \includegraphics[width=0.9\textwidth]{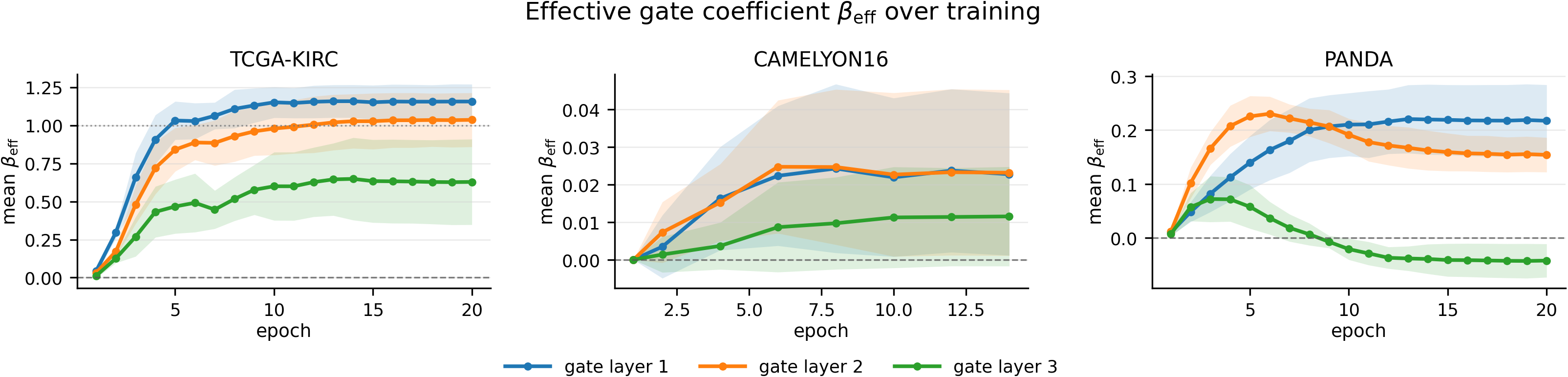}
  \caption{\textbf{Effective gate trajectories for representative datasets.} We plot $\beta_{\mathrm{eff}}=\delta\tanh(g)$, where $g$ is the learned gate logit, for the three gate-active Gated SRP layers. Curves show the mean across five global seeds, where each seed-level point averages the logged training-batch $\beta_{\mathrm{eff}}$ values within an epoch. Shaded regions denote $\pm1$ standard deviation across seeds.}
  \label{fig:beta-trajectory-main}
\end{figure*}

\resultpara{Gate trajectory} We track the effective signed coefficient $\beta_{\mathrm{eff}}$ during training to study gate behaviour. Each value is averaged over heads and patch tokens in the current batch. Figure~\ref{fig:beta-trajectory-main} shows trajectories per epoch averaged over five global seeds. We include TCGA-KIRC, CAMELYON16, and PANDA because they represent three regimes. In TCGA-KIRC, the first two active layers rise above $\beta_{\mathrm{eff}}=1$, indicating correction stronger than pure projection of the local common component. This is consistent with the possibility that survival prediction benefits from stronger suppression of homogeneous local context, although further analysis is needed to connect gate values to specific prognostic tissue patterns. In CAMELYON16, all gates stay near zero, which is expected for a saturated metastasis detection task where the NA baseline already performs strongly, and Gated SRP can remain close to identity. For PANDA, the first two layers learn modest positive corrections, and the third layer becomes slightly negative, suggesting that early layers reduce redundant local tissue context, whereas a deeper layer can preserve or reintroduce local grade context for ordered prostate grading. Full trajectories for the remaining datasets are provided in the appendix.

\subsection{Qualitative Results}
\label{sec:qualitative-result}

\begin{figure}[ht]
  \centering
  \includegraphics[width=0.9\textwidth]{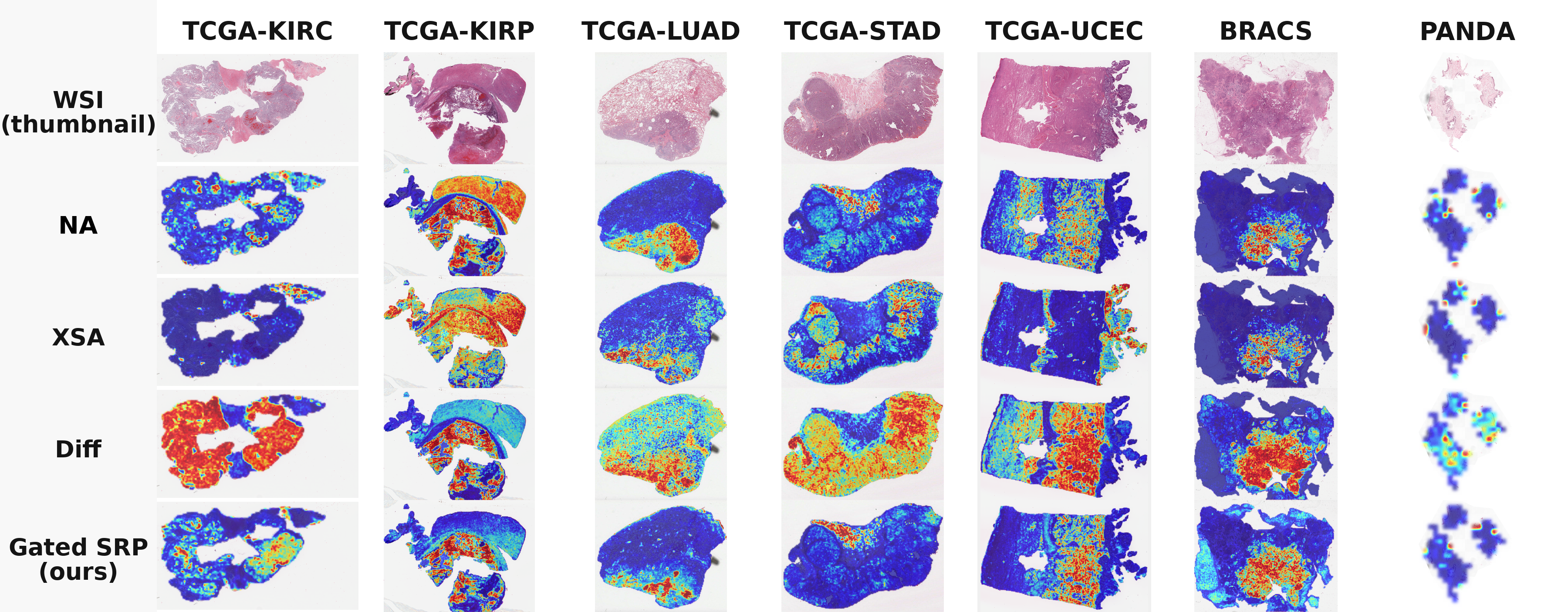}
  \caption{\textbf{Qualitative comparison of attention maps.} One representative WSI from some datasets. Each heatmap overlays the final layer CLS-to-patch attention score on the WSI thumbnail and is normalized within each slide and method for visualization. For Diff, the final attention operator is signed, $A_1-\lambda A_2$. We report its positive support for comparison. Warmer colours consistently indicate higher attention.}

  \label{fig:heatmap-grid}
\end{figure}

For the qualitative study, a random selection of 10 WSIs per dataset and their attention maps were reviewed by a board-certified pathologist. Figure~\ref{fig:heatmap-grid} shows one representative WSI from some datasets. The pathologist notably studied how attention maps focused on the lesion, if the higher attention areas within the lesion correlated with known pathology features, and the degree of background attention on areas of less known biological and pathological value. Attention maps differed between models based on the task and the dataset. Gated SRP and the base NA showed mostly similar behavior, while XSA slightly diverged on these aspects. Diff was noticeably divergent compared to the other models.

Overall, in the reviewed examples, Gated SRP attention maps appeared to place higher attention on lesion regions and showed better visual agreement with pathologist-identified tissue features. This observation was consistent for overall survival prediction tasks from the TCGA-KIRC, KIRP, LUAD, and STAD datasets. The only exception was with UCEC. Similar to the base NA, Gated SRP focuses on the underlying normal tissue that the cancer was invading, possibly due to the training signal pushing Gated SRP back to the base attention. For classification tasks, Gated SRP specifically identified the class lesion at higher rates than the other models. These observations provide qualitative examples that are consistent with the quantitative trends.

%% file: sections/Ablation.tex
\subsection{Ablation Studies}
\label{sec:exp-ablations}

Our ablations focus on survival analysis tasks using three TCGA datasets (KIRP, LUAD, STAD) \cite{NCI_TCGA}. For classification tasks, we use KGH \cite{adpv2}, PANDA \cite{panda}, and BRACS \cite{bracs}. All ablations use the same embedded patch features, data splits, optimizer, training schedule, and five global seeds as the main experiments. Each ablation changes one design choice while keeping the remaining Gated SRP configuration fixed.

\resultpara{Ablation on fixed projection strength}We replace the adaptive gate with fixed scalar projection strengths. This ablation keeps the same local spatial redundancy axis and changes only the correction strength. As shown in Table~\ref{tab:ablation-fixed-beta}, none of the fixed projections is uniformly reliable. In contrast, Gated SRP learns the correction coefficient from local neighbourhood geometry and head diagnostics, producing a more stable improvement pattern than the base NA. These results support the use of an adaptive local gate in our evaluated setting.

\begin{table}[ht]
\centering
\caption{\textbf{Fixed projection strength compared with a learned local gate.} All rows use the same local spatial redundancy axis and differ only in the coefficient used to correct the projected component. $\beta=0$ recovers the base attention, $\beta=1$ removes the local component, $\beta=2$ reflects it, and $\beta=-1$ amplifies it. The final row uses the proposed learned local gate.}
\label{tab:ablation-fixed-beta}
{\scriptsize
\setlength{\tabcolsep}{2.0pt}
\renewcommand{\arraystretch}{1.04}
\resizebox{\textwidth}{!}{%
\begin{tabular}{l*{13}{c}}
\toprule
\textbf{Config} & \textbf{KIRP} & \textbf{LUAD} & \textbf{STAD} & \multicolumn{3}{c}{\textbf{KGH}} & \multicolumn{4}{c}{\textbf{PANDA}} & \multicolumn{3}{c}{\textbf{BRACS}} \\
\cmidrule(lr){2-4}\cmidrule(lr){5-7}\cmidrule(lr){8-11}\cmidrule(lr){12-14}
& \textit{C} & \textit{C} & \textit{C} & \textit{F1} & \textit{Acc} & \textit{AUC} & $\mathit{\kappa_q}$ & \textit{F1} & \textit{Acc} & \textit{AUC} & \textit{F1} & \textit{Acc} & \textit{AUC} \\
\midrule
Base Attention ($\beta=0$) & 0.7247 & 0.5513 & 0.5910 & 0.8297 & 0.8337 & 0.9665 & 0.8822 & 0.6861 & 0.7374 & 0.9313 & 0.4179 & 0.5439 & \textbf{0.8472} \\
Projection ($\beta=1$) & \textcolor[HTML]{B3261E}{0.6863} & \textcolor[HTML]{B3261E}{0.5319} & \textcolor[HTML]{B3261E}{0.5571} & 0.8299 & 0.8313 & \textcolor[HTML]{B3261E}{0.9628} & 0.8829 & 0.6763 & 0.7290 & \textcolor[HTML]{137333}{\textbf{0.9356}} & \textcolor[HTML]{137333}{0.4323} & \textcolor[HTML]{137333}{0.5585} & \textcolor[HTML]{B3261E}{0.8396} \\
Reflection ($\beta=2$) & \textcolor[HTML]{B3261E}{0.7051} & \textcolor[HTML]{B3261E}{0.5290} & \textcolor[HTML]{B3261E}{0.5669} & \textcolor[HTML]{B3261E}{0.8217} & \textcolor[HTML]{B3261E}{0.8229} & \textcolor[HTML]{B3261E}{0.9625} & 0.8819 & 0.6804 & 0.7335 & 0.9308 & \textcolor[HTML]{137333}{0.4224} & \textcolor[HTML]{137333}{0.5585} & \textcolor[HTML]{B3261E}{0.8401} \\
Amplification ($\beta=-1$) & \textcolor[HTML]{B3261E}{0.6854} & \textcolor[HTML]{137333}{0.5655} & \textcolor[HTML]{B3261E}{0.5692} & 0.8251 & 0.8277 & 0.9676 & \textcolor[HTML]{B3261E}{0.8771} & 0.6769 & 0.7346 & \textcolor[HTML]{B3261E}{0.9243} & \textcolor[HTML]{137333}{\textbf{0.4442}} & \textcolor[HTML]{137333}{0.5585} & \textcolor[HTML]{B3261E}{0.8370} \\
\midrule
\textbf{Gated SRP} & \textcolor[HTML]{137333}{\textbf{0.7648}} & \textcolor[HTML]{137333}{\textbf{0.5832}} & \textcolor[HTML]{137333}{\textbf{0.6171}} & \textcolor[HTML]{137333}{\textbf{0.8544}} & \textcolor[HTML]{137333}{\textbf{0.8566}} & \textcolor[HTML]{137333}{\textbf{0.9695}} & \textbf{0.8842} & \textcolor[HTML]{137333}{\textbf{0.6932}} & \textcolor[HTML]{137333}{\textbf{0.7461}} & 0.9292 & \textcolor[HTML]{137333}{0.4424} & \textcolor[HTML]{137333}{\textbf{0.5603}} & 0.8454 \\
\bottomrule
\end{tabular}}}
\end{table}

\resultpara{Ablation on gate range}To confirm the necessity of a signed gate range, we replace the tanh gate with a sigmoid gate while keeping the $\delta$ and gate hidden dimension fixed. The signed tanh gate allows the correction coefficient to be negative or positive, while the sigmoid gate restricts it to nonnegative values. Table~\ref{tab:ablation-gate-range} suggests that the signed gate gives stronger overall mean performance in this setting. This result indicates that a one-sided nonnegative correction is too restrictive for local spatial redundancy correction. Instead, the stronger performance of the signed gate suggests that the model sometimes benefits from preserving or increasing the local component. This is consistent with the pathology setting, where nearby patches may either provide repeated tissue context or contain useful local evidence depending on the tissue region. The signed formulation is more flexible and avoids forcing every local common component only to be subtracted.

\begin{table}[ht]
\centering
\caption{\textbf{Gate range ablation.} We compare the signed tanh gate with a nonnegative sigmoid gate while keeping each dataset's selected $\delta$ and gate hidden dimension fixed. The signed range $[-\delta,\delta]$ allows both positive subtraction and negative reinforcement of the local common component. The nonnegative range $[0,\delta]$ removes the negative branch and only permits nonnegative correction coefficients.}
\label{tab:ablation-gate-range}
{\scriptsize
\setlength{\tabcolsep}{2.0pt}
\renewcommand{\arraystretch}{1.04}
\resizebox{\textwidth}{!}{%
\begin{tabular}{l*{13}{c}}
\toprule
\textbf{Config} & \textbf{KIRP} & \textbf{LUAD} & \textbf{STAD} & \multicolumn{3}{c}{\textbf{KGH}} & \multicolumn{4}{c}{\textbf{PANDA}} & \multicolumn{3}{c}{\textbf{BRACS}} \\
\cmidrule(lr){2-4}\cmidrule(lr){5-7}\cmidrule(lr){8-11}\cmidrule(lr){12-14}
& \textit{C} & \textit{C} & \textit{C} & \textit{F1} & \textit{Acc} & \textit{AUC} & $\mathit{\kappa_q}$ & \textit{F1} & \textit{Acc} & \textit{AUC} & \textit{F1} & \textit{Acc} & \textit{AUC} \\
\midrule
Sigmoid $[0,\delta]$ & 0.7269 & 0.5512 & 0.5871 & 0.8338 & 0.8349 & 0.9635 & 0.8821 & 0.6761 & 0.7336 & 0.9236 & 0.4148 & 0.5421 & \textbf{0.8464} \\
\textbf{Signed tanh $[-\delta,\delta]$} & \textbf{0.7648} & \textbf{0.5832} & \textbf{0.6171} & \textbf{0.8544} & \textbf{0.8566} & \textbf{0.9695} & \textbf{0.8842} & \textbf{0.6932} & \textbf{0.7461} & \textbf{0.9292} & \textbf{0.4424} & \textbf{0.5603} & 0.8454 \\
\bottomrule
\end{tabular}}}
\end{table}



\resultpara{Ablation on gate input gradients}We compare detached gate inputs with live gate inputs to determine whether the local spatial reference should act as measured geometry or be optimized through the gate input path. In the detached setting, the local reference and diagnostic features are computed from the patch stream but do not send additional gradients back into that stream. The gate is still trained through its effect on the corrected attention output. In the live setting, gradients from the diagnostic construction also update the patch stream. Table~\ref{tab:ablationGateInputGradient} shows that detached inputs give stronger results on most metrics. This suggests that the local spatial reference is more reliable when it is used as a measured guide rather than as an additional optimization path. Thus, we use stop gradient in the default design, so the gate learns how strongly to apply the correction while the diagnostic reference remains stable.

\begin{table}[ht]
\centering
\caption{\textbf{Gate input gradient ablation.} Detached inputs treat the local spatial reference and gate diagnostics as measured geometry. Live inputs allow gradients from those diagnostics to propagate into the patch stream.}
\label{tab:ablationGateInputGradient}
{\scriptsize
\setlength{\tabcolsep}{2.0pt}
\renewcommand{\arraystretch}{1.04}
\resizebox{\textwidth}{!}{%
\begin{tabular}{l*{13}{c}}
\toprule
\textbf{Gradient path} & \textbf{KIRP} & \textbf{LUAD} & \textbf{STAD} & \multicolumn{3}{c}{\textbf{KGH}} & \multicolumn{4}{c}{\textbf{PANDA}} & \multicolumn{3}{c}{\textbf{BRACS}} \\
\cmidrule(lr){2-4}\cmidrule(lr){5-7}\cmidrule(lr){8-11}\cmidrule(lr){12-14}
& \textit{C} & \textit{C} & \textit{C} & \textit{F1} & \textit{Acc} & \textit{AUC} & $\mathit{\kappa_q}$ & \textit{F1} & \textit{Acc} & \textit{AUC} & \textit{F1} & \textit{Acc} & \textit{AUC} \\
\midrule
Live inputs & 0.7177 & 0.5349 & 0.5902 & 0.8386 & 0.8410 & 0.9621 & 0.8795 & 0.6793 & 0.7333 & 0.9273 & 0.4166 & 0.5439 & \textbf{0.8475} \\
\textbf{Detached inputs} & \textbf{0.7648} & \textbf{0.5832} & \textbf{0.6171} & \textbf{0.8544} & \textbf{0.8566} & \textbf{0.9695} & \textbf{0.8842} & \textbf{0.6932} & \textbf{0.7461} & \textbf{0.9292} & \textbf{0.4424} & \textbf{0.5603} & 0.8454 \\
\bottomrule
\end{tabular}}}
\end{table}

\resultpara{Ablation on patch encoder}We test whether Gated SRP depends on the frozen patch encoder used to construct the slide tokens. Besides UNI-v2~\cite{UNI}, we evaluate MedSigLIP-448 \cite{sellergren2025medgemma}, a medical image-text encoder with a larger input field, and ViT-B/16 \cite{vit}, a generic vision transformer encoder pretrained on ImageNet. Table~\ref{tab:ablation-encoder} shows that Gated SRP provides improvements on many metrics across all three feature spaces. The most consistent improvements appear with UNI-v2, which suggests that a pathology-specific visual feature space provides a more reliable local reference for spatial redundancy correction. MedSigLIP-448 also improves all TCGA survival cohorts and several classification metrics, showing that medical image-text pretraining can still provide useful features for Gated SRP. ViT-B/16 is less aligned with pathology, so its performance remains weaker than other encoders, but still benefits from Gated SRP on several metrics. Overall, Gated SRP is compatible with multiple encoders, and its benefit is strongest when the patch features are more aligned with histological morphology.

\begin{table}[ht]
\centering
\caption{\textbf{Patch encoder ablation.} Base NA and Gated SRP are compared using frozen patch features from UNI-v2, MedSigLIP-448, and ViT-B/16. All image tiles are extracted at $20\times$ optical magnification. Tile sizes are $256 \times 256$ for UNI-v2, $448 \times 448$ for MedSigLIP-448, and $224 \times 224$ for ViT-B/16.}
\label{tab:ablation-encoder}
{\scriptsize
\setlength{\tabcolsep}{2.0pt}
\renewcommand{\arraystretch}{1.04}
\resizebox{\textwidth}{!}{%
\begin{tabular}{ll*{13}{c}}
\toprule
\textbf{Encoder} & \textbf{Attn.} & \textbf{KIRP} & \textbf{LUAD} & \textbf{STAD} & \multicolumn{3}{c}{\textbf{KGH}} & \multicolumn{4}{c}{\textbf{PANDA}} & \multicolumn{3}{c}{\textbf{BRACS}} \\
\cmidrule(lr){3-5}\cmidrule(lr){6-8}\cmidrule(lr){9-12}\cmidrule(lr){13-15}
& & \textit{C} & \textit{C} & \textit{C} & \textit{F1} & \textit{Acc} & \textit{AUC} & $\mathit{\kappa_q}$ & \textit{F1} & \textit{Acc} & \textit{AUC} & \textit{F1} & \textit{Acc} & \textit{AUC} \\
\midrule
\multirow{2}{*}{ViT-B/16} & NA & 0.6263 & 0.5679 & 0.5871 & 0.7919 & 0.7964 & 0.9532 & \textbf{0.7870} & 0.5545 & 0.6181 & 0.8851 & 0.3925 & 0.5090 & \textbf{0.8095} \\
& Gated SRP & \textbf{0.6765} & \textbf{0.6043} & \textbf{0.6052} & \textbf{0.7966} & \textbf{0.8000} & \textbf{0.9537} & 0.7851 & \textbf{0.5649} & \textbf{0.6266} & \textbf{0.8878} & \textbf{0.4051} & \textbf{0.5146} & 0.8061 \\
\midrule
\multirow{2}{*}{MSL-448} & NA & 0.6400 & 0.5223 & 0.5744 & \textbf{0.8010} & \textbf{0.8048} & \textbf{0.9572} & 0.8308 & 0.6161 & 0.6753 & 0.9120 & 0.4561 & 0.5676 & \textbf{0.8558} \\
& Gated SRP & \textbf{0.6943} & \textbf{0.5438} & \textbf{0.6057} & 0.7958 & 0.8000 & 0.9539 & \textbf{0.8348} & \textbf{0.6193} & \textbf{0.6817} & \textbf{0.9133} & \textbf{0.4645} & \textbf{0.5768} & 0.8540 \\
\midrule
\multirow{2}{*}{UNI-v2} & NA & 0.7247 & 0.5513 & 0.5910 & 0.8297 & 0.8337 & 0.9665 & 0.8822 & 0.6861 & 0.7374 & \textbf{0.9313} & 0.4179 & 0.5439 & \textbf{0.8472} \\
& Gated SRP & \textbf{0.7648} & \textbf{0.5832} & \textbf{0.6171} & \textbf{0.8544} & \textbf{0.8566} & \textbf{0.9695} & \textbf{0.8842} & \textbf{0.6932} & \textbf{0.7461} & 0.9292 & \textbf{0.4424} & \textbf{0.5603} & 0.8454 \\
\bottomrule
\end{tabular}}}
\end{table}


\resultpara{Ablation on gate factorization}To understand which signals drive the gate, we separate its information sources. The token-only row measures the effect of local geometry and neighbourhood counts. The head-only row uses attention head diagnostics without token diagnostics. Removing the learned bias tests whether the gate needs a task-specific prior. In Table~\ref{tab:ablation-gate-factorization}, the full gate is mostly preferred. The token-only gate is never the best configuration, showing that local geometry alone is not sufficient for reliable correction. The head-only gate performs best on BRACS, while the gate without bias is strongest on LUAD. These exceptions suggest that some datasets benefit from simpler gate behaviour, but the full gate provides the most stable design across tasks by combining local tissue statistics, head-level responses, and a learned offset.

\begin{table}[ht]
\centering
\caption{\textbf{Gate factorization ablation.} We isolate the full Gated SRP gate, a token-only gate, a head-only gate, and a gate without the learned bias term.}
\label{tab:ablation-gate-factorization}
{\scriptsize
\setlength{\tabcolsep}{2.0pt}
\renewcommand{\arraystretch}{1.04}
\resizebox{\textwidth}{!}{%
\begin{tabular}{l*{13}{c}}
\toprule
\textbf{Gate form} & \textbf{KIRP} & \textbf{LUAD} & \textbf{STAD} & \multicolumn{3}{c}{\textbf{KGH}} & \multicolumn{4}{c}{\textbf{PANDA}} & \multicolumn{3}{c}{\textbf{BRACS}} \\
\cmidrule(lr){2-4}\cmidrule(lr){5-7}\cmidrule(lr){8-11}\cmidrule(lr){12-14}
& \textit{C} & \textit{C} & \textit{C} & \textit{F1} & \textit{Acc} & \textit{AUC} & $\mathit{\kappa_q}$ & \textit{F1} & \textit{Acc} & \textit{AUC} & \textit{F1} & \textit{Acc} & \textit{AUC} \\
\midrule
Token only & 0.7038 & 0.5471 & 0.5722 & 0.8301 & 0.8325 & 0.9602 & 0.8793 & 0.6641 & 0.7248 & 0.9280 & 0.4312 & 0.5530 & 0.8458 \\
Head only & 0.7171 & 0.5815 & 0.5923 & 0.8353 & 0.8373 & 0.9610 & 0.8774 & 0.6687 & 0.7261 & 0.9218 & \textbf{0.4471} & \textbf{0.5640} & \textbf{0.8499} \\
No bias & 0.7087 & \textbf{0.5922} & 0.5947 & 0.8111 & 0.8157 & 0.9621 & 0.8784 & 0.6842 & 0.7335 & \textbf{0.9342} & 0.4146 & 0.5512 & 0.8481 \\
\midrule
\textbf{Full gate} & \textbf{0.7648} & 0.5832 & \textbf{0.6171} & \textbf{0.8544} & \textbf{0.8566} & \textbf{0.9695} & \textbf{0.8842} & \textbf{0.6932} & \textbf{0.7461} & 0.9292 & 0.4424 & 0.5603 & 0.8454 \\
\bottomrule
\end{tabular}}}
\end{table}


\resultpara{Ablation on gate initialization}Initialization controls whether the model starts from the standard self-attention solution or from an immediate projection correction. Zero initialization isolates the effect of learning the correction during training. Xavier \cite{xavier} and Kaiming \cite{kaiming} initialization introduce a nonzero correction at the first step to probe optimization stability. Table~\ref{tab:ablation-gate-initialization} shows that zero initialization is favourable in most scenarios, while Xavier and Kaiming are sometimes preferred, suggesting that some tasks may benefit from an early correction signal. In most of the cases, Gated SRP is more reliable when it starts from the identity solution and learns the projection correction gradually during training. 
\begin{table}[ht]
\centering
\caption{\textbf{Gate initialization ablation.} Zero initialization starts from the Standard SA attention contribution because the initial projection coefficient is zero. Xavier and Kaiming initialization test whether a non-identity starting gate changes optimization.}
\label{tab:ablation-gate-initialization}
{\scriptsize
\setlength{\tabcolsep}{2.0pt}
\renewcommand{\arraystretch}{1.04}
\resizebox{\textwidth}{!}{%
\begin{tabular}{l*{13}{c}}
\toprule
\textbf{Init.} & \textbf{KIRP} & \textbf{LUAD} & \textbf{STAD} & \multicolumn{3}{c}{\textbf{KGH}} & \multicolumn{4}{c}{\textbf{PANDA}} & \multicolumn{3}{c}{\textbf{BRACS}} \\
\cmidrule(lr){2-4}\cmidrule(lr){5-7}\cmidrule(lr){8-11}\cmidrule(lr){12-14}
& \textit{C} & \textit{C} & \textit{C} & \textit{F1} & \textit{Acc} & \textit{AUC} & $\mathit{\kappa_q}$ & \textit{F1} & \textit{Acc} & \textit{AUC} & \textit{F1} & \textit{Acc} & \textit{AUC} \\
\midrule
Xavier & 0.7271 & 0.5805 & 0.5799 & 0.8369 & 0.8373 & 0.9643 & \textbf{0.8857} & 0.6900 & 0.7423 & \textbf{0.9358} & 0.4466 & \textbf{0.5658} & \textbf{0.8565} \\
Kaiming & 0.6550 & 0.5587 & 0.5775 & 0.8343 & 0.8361 & 0.9646 & 0.8758 & 0.6684 & 0.7248 & 0.9295 & \textbf{0.4503} & 0.5530 & 0.8527 \\
\midrule
\textbf{Zero} & \textbf{0.7648} & \textbf{0.5832} & \textbf{0.6171} & \textbf{0.8544} & \textbf{0.8566} & \textbf{0.9695} & 0.8842 & \textbf{0.6932} & \textbf{0.7461} & 0.9292 & 0.4424 & 0.5603 & 0.8454 \\
\bottomrule
\end{tabular}}}
\end{table}

\resultpara{Ablation on architecture choice}Our main experiments use the TransMIL-style NA backbone because dense self-attention is computationally expensive for large WSIs. To test whether Gated SRP also works in a standard MHSA setting, we evaluate it on a ViT \cite{vit} model using ADP~\cite{adp} and PANDA~\cite{panda}, where the image sizes are small enough to make dense attention feasible. Table~\ref{tab:ablation-dense-attention-setting} shows that Gated SRP improves all reported metrics on both datasets. The gains are modest but consistent, indicating that the proposed correction is not tied to the Nystr\"om approximation. This supports Gated SRP as a general attention correction, while NA remains the practical backbone for larger slide-level cohorts.
\begin{table}[ht]
\centering
\caption{\textbf{Architecture choice ablation.} ADP uses patch-level multilabel classification and reports mAP, F1, accuracy, and AUC. PANDA uses slide-level grading and additionally reports quadratic weighted $\kappa_q$. Both methods use the ViT training protocol and differ only in the attention update.}
\label{tab:ablation-dense-attention-setting}
{\scriptsize
\setlength{\tabcolsep}{2.2pt}
\renewcommand{\arraystretch}{1.04}
\begin{tabular}{@{}lcccccccc@{}}
\toprule
\textbf{Method} & \multicolumn{4}{c}{\textbf{ADP}} & \multicolumn{4}{c}{\textbf{PANDA}} \\
\cmidrule(lr){2-5}\cmidrule(lr){6-9}
& \textit{mAP} & \textit{F1} & \textit{Acc} & \textit{AUC} & $\mathit{\kappa_q}$ & \textit{F1} & \textit{Acc} & \textit{AUC} \\
\midrule
MHSA & 0.7768 & 0.6997 & 0.9360 & 0.9424 & 0.8801 & 0.6663 & 0.7264 & 0.9331 \\
\textbf{Gated SRP} & \textbf{0.7813} & \textbf{ 0.7031} & \textbf{0.9372} & \textbf{0.9439} & \textbf{0.8820} & \textbf{0.6721} & \textbf{0.7352} & \textbf{0.9351} \\
\bottomrule
\end{tabular}}
\end{table}

%% file: sections/conclusion.tex
\section{Conclusion}
We presented Gated SRP, a lightweight post-attention correction module for pathology transformers. Motivated by the local spatial redundancy of WSIs, Gated SRP corrects each patch-token attention update along a learned local tissue-neighbourhood axis while preserving the original attention routing, classifier, and loss. Across five TCGA survival cohorts and five slide-level classification datasets, Gated SRP achieves favorable mean performance compared with the base NA attention while adding negligible parameter overhead. The ablation studies support the importance of the spatial axis, adaptive gate, and signed correction range. These results suggest that explicitly correcting neighbourhood-aligned redundancy is a promising and lightweight way to adapt transformer attention to pathology images.

\resultpara{Limitations and Future Work}Although Gated SRP is theoretically compatible with any self-attention block, the test of its effect on other transformer architectures, attention variants, and multi-scale WSI designs is still narrow. In addition, all main WSI evaluations use supervised downstream training with frozen patch features, while the ADP architecture choice ablation is the exception and trains from raw RGB patches. Given the increasing importance of self-supervised pretraining in pathology, future work should study whether Gated SRP can improve representation learning during large-scale WSI pretraining. Further directions include testing the learned gate across different magnifications, analyzing its relationship with histological structures and microenvironment patterns, and validating whether the corrected attention maps provide more biologically meaningful evidence for prognostic modelling.

%% file: sections/supplementary.tex
\section{Dataset Details}
\label{dataset_details}
\paragraph{CAMELYON16.}
CAMELYON16 \cite{cam16} is a lymph-node metastasis benchmark for breast cancer. We use it to perform binary WSI-level metastasis classification, where tumour slides are positive and normal slides are negative. The original challenge provides $270$ labelled training WSIs and $129$ official test WSIs. Our experiments use all $270$ WSIs, consisting of $111$ tumour and $159$ normal slides. The $129$ official test WSIs are not used because our protocol builds train, validation, and test splits from the same labelled feature inventory shared by all experiments. All splits are generated with fixed global random seeds to ensure reproducibility across methods.

\paragraph{CAMELYON17.}
CAMELYON17 \cite{cam17} extends lymph-node metastasis evaluation to multiple lymph-node slides per patient. The tested task is a four-class slide-level metastasis stage classification, using the labels negative, isolated tumour cells (ITC), micro-metastasis, and macro-metastasis. The public challenge contains $1,000$ WSIs, with $500$ labelled training slides and $500$ unlabeled test slides. We use $499$ valid labelled training WSIs after preprocessing. The $500$ unlabeled challenge test slides are not used because they do not provide slide labels for the supervised classification protocol. There are $318$ negative, $36$ ITC, $59$ micro-metastasis, and $86$ macro-metastasis slides. We split the data by patient rather than by slide to prevent information leakage, ensuring that all slides from the same patient appear in only one of the training, validation, or test sets.

\paragraph{ADP.}
ADP \cite{adp} is used only in the architecture choice ablation as a patch-level multilabel histological tissue-type classification task. The Release1 flat table contains $17{,}668$ RGB patches at $272 \times 272$ pixels and $1\,\mu$m/pixel, with $43$ hierarchical multi-hot labels. We use the official patch-level train, validation, and test split files with $14{,}134$, $1{,}767$, and $1{,}767$ patches, respectively. Five level-3 labels with zero training positives are pruned from the model output, leaving $38$ evaluated labels. The ADP experiment trains a raw-RGB ViT-S/16 patch classifier with $12$ transformer layers, embedding width $384$, $6$ attention heads, and a $17 \times 17$ patch-token grid. The primary metric is macro-average precision. Because ADP is patch-level and its official split is not a slide-level WSI split, it is not used as a main WSI evaluation.

\paragraph{ADPv2-KGH.}
KGH \cite{adpv2} is used as a colorectal polyp WSI classification task. It is an in-house collection of $1,037$ WSIs in the ADPv2 work for distribution analysis and potential biomarker discovery. The tested task is four-class disease-subtype classification over hyperplastic polyp (\texttt{CP\_HP}), sessile serrated lesion (\texttt{CP\_SSL}), tubular adenoma (\texttt{CP\_TA}), and tubulovillous adenoma (\texttt{CP\_TVA}). The raw local tree contains $1,037$ WSIs: $837$ disease-subtype WSIs and $200$ normal WSIs. The $200$ normal slides are not included. Among the four target subtypes, all $837$ slides are used, including $212$ \texttt{CP\_HP}, $201$ \texttt{CP\_SSL}, $207$ \texttt{CP\_TA}, and $217$ \texttt{CP\_TVA} slides.

\paragraph{PANDA.}
PANDA \citep{panda} is a large public prostate biopsy grading benchmark released for the Prostate cANcer graDe Assessment challenge. The tested task is a six-class WSI-level ISUP grade classification over grades $0$ to $5$, and the main table additionally reports quadratic-weighted Cohen's kappa because grade ordering is clinically meaningful. The local label file contains $10,616$ labelled H\&E biopsy WSIs. We use $10,615$ valid slides after preprocessing. The usable label distribution is $2,891$, $2,666$, $1,343$, $1,242$, $1,249$, and $1,224$ slides for grades $0-5$, respectively.

\paragraph{BRACS.}
BRACS \cite{bracs} contains H\&E breast pathology WSIs and ROI annotations for breast lesion subtyping. The tested task is seven-class WSI-level lesion classification using the WSI labels, not the ROI-level labels. The public dataset includes $547$ WSIs and $4,539$ ROIs. We use $546$ valid WSI-level slides for training and evaluation. One WSI is excluded after preprocessing. The seven target classes are normal (N), pathological benign (PB), usual ductal hyperplasia (UDH), flat epithelial atypia (FEA), atypical ductal hyperplasia (ADH), ductal carcinoma in situ (DCIS), and invasive carcinoma (IC), which includes $44$ N, $147$ PB, $73$ UDH, $41$ FEA, $48$ ADH, $61$ DCIS, and $132$ IC.

\paragraph{TCGA survival cohorts.}
The survival benchmark uses TCGA-KIRC, KIRP, LUAD, STAD, and UCEC \cite{NCI_TCGA}. The tested task is the ranking of overall survival (OS) risk. Each WSI is processed as a slide bag, but splitting and evaluation are performed at the case level. All slides from the same case stay in the same split. Multi-slide cases are aggregated into a single case risk score, and performance is measured by the case C-index. We start from the matched label table and remove rows with non-positive survival times before training. The final valid inventories are $511$ KIRC cases ($517$ slides), $272$ KIRP cases ($296$ slides), $465$ LUAD cases ($528$ slides), $383$ STAD cases ($409$ slides), and $249$ UCEC cases ($275$ slides). Event counts are $171$, $42$, $163$, $156$, and $34$ for KIRC, KIRP, LUAD, STAD, and UCEC, respectively.

\section{Implementation Details}

All WSI experiments use the same frozen features across datasets. Patches are extracted with AtlasPatch~\cite{atlaspatch} at $20\times$ magnification and embedded with UNI-v2~\cite{UNI}. The patch-encoder ablation uses the same protocol but replaces the UNI-v2 features with MedSigLIP-448 or ViT-B/16 features. The ADP experiment is the exception. It uses the raw RGB patch images rather than WSI feature bags. The PANDA experiment uses the same UNI-v2 PANDA slide features as the WSI experiments, but routes them through a full-attention set-style ViT instead of the TransMIL-style Nystr\"om aggregator.

\paragraph{Model architecture}
To handle exploding memory consumption issues when using dense ViT attention for WSIs, we use a TransMIL-style~\cite{transmil} slide aggregator with Nystr\"om attention and spatial neighbourhoods constructed from patch coordinates. This architecture is also used for PANDA in the main experiments. Our aggregator follows TransMIL's overall recipe (CLS token + Nystr\"om
attention + PPEG, square-replication padding, PPEG inserted once after
the first attention block) but is not a drop-in re-implementation. We
deviate in (i) \emph{embedding width} ($384$ vs $512$) and \emph{head
count} ($6$ vs $8$), chosen to match the upstream ViT-S used for
diagnostic continuity with our early patch-level analysis that is not included in this manuscript; (ii) \emph{depth}
($4$ vs $2$ TransLayers), for finer per-layer training trajectory diagnostics;
(iii) \emph{block topology}, where each of our TransLayers is a standard
pre-norm transformer block with both an attention sub-block and
an MLP sub-block (\textsc{LayerNorm}~$\rightarrow$~attention~$\rightarrow$~residual~$\rightarrow$~\textsc{LayerNorm}~$\rightarrow$~MLP~$\rightarrow$~residual), whereas official TransMIL
retains only the attention sub-block; (iv) \emph{Nystr\"om landmarks}
($m{=}64$ vs $m{=}\mathrm{dim}/2{=}256$), the
\texttt{lucidrains/nystrom-attention} package default, chosen for
memory economy; (v) the \emph{value-residual depthwise convolution}
inside Nystr\"om (\texttt{residual=True} in the
\texttt{lucidrains/nystrom-attention} import path used by official
TransMIL) is \emph{disabled} here, because a 1-D depthwise convolution
over an unordered token sequence is not geometrically meaningful in
the MIL setting; and (vi) \emph{regularization choices} -- we add
stochastic depth (linear schedule $0 \to 0.1$ across depth) and use
truncated-normal initialisation ($\mathrm{std}{=}0.02$) for all linear
and convolutional weights, which are commonly used in standard transformer models but not in official TransMIL.

For the architecture choice ablation, ADP uses a raw-RGB ViT-S/16 patch classifier. Each $272 \times 272$ patch is split into a $17 \times 17$ grid of $16 \times 16$ visual tokens and a \textsc{cls} token. The model uses $12$ transformer blocks with width $384$, $6$ attention heads, an MLP ratio of $4$, learned absolute positional embeddings, and stochastic depth with a maximum drop path of $0.1$. The ADP Gated SRP arm replaces dense MHSA blocks with a patch-level Gated SRP attention backend; the local reference is built on a fixed $17 \times 17$ token grid.

For PANDA, we use the \texttt{vit4} slide model rather than the main TransMIL-style PANDA model. Each WSI is represented by native-length UNI-v2 patch features of dimension $1536$ and their level-$0$ patch coordinates. A linear projection maps each patch feature to width $384$, a learned \textsc{cls} token is prepended, and $4$ full-attention transformer blocks with $6$ heads and an MLP ratio of $4$ aggregate the variable-length slide bag under an attention mask. No learned positional encoding is used in the reported \texttt{vit4} setting. For Gated SRP, the model uses the coordinate-derived $3 \times 3$ neighbourhood graph, equivalent to a valid $8$-neighbour local context, with the local reference target set to \texttt{knn8}.

\paragraph{Training objectives and model selection.}
For slide-level classification, we use standard unweighted cross entropy over the class labels of each dataset. The checkpoint used for final test evaluation is selected by validation macro F1 for CAMELYON16, CAMELYON17, KGH, and BRACS. PANDA is also trained with unweighted cross entropy, but its checkpoint is selected by validation quadratic weighted Cohen's kappa to match the ordinal grading objective used by the PANDA challenge. We do not use class weighted losses, oversampling, or weighted random sampling. Instead, class imbalance is handled by stratified splitting and by reporting macro or ordinal metrics, including macro F1, macro AUC, and $\kappa_q$.

For ADP, the architecture choice ablation task is patch-level multilabel classification over the $38$ retained labels. We train with binary cross entropy with logits and select the checkpoint by validation macro mAP. Training patches use random horizontal and vertical flips, discrete $90^\circ$ rotations, mild colour jitter, and ImageNet normalization; validation and test patches use deterministic tensor conversion and the same normalization. ADP F1 and accuracy are computed with the standard multilabel threshold at logit $0$, and ADP AUC is the macro average over labels with both positive and negative samples in the evaluated split.

For TCGA survival, we train a discrete time survival model with $4$ time bins. The model outputs one logit for each time bin, and the sigmoid of each logit is interpreted as the conditional hazard for that bin. Bin edges are estimated from uncensored training cases for each cohort and seed. If a split has too few event times, all training survival times are used so that the bins remain well defined. The loss is the discrete time negative log likelihood. For an uncensored event in bin $y$, the likelihood includes survival through the bins before $y$ and event occurrence in bin $y$. For a censored case in bin $y$, the likelihood includes survival through bin $y$. Thus, censored samples contribute information up to the censoring time and are not treated as observed events. During evaluation, predicted hazards are converted to a scalar risk score by taking the negative sum of the predicted discrete survival curve. Larger risk values therefore correspond to earlier predicted events.

TCGA evaluation is case-level. Each WSI is forwarded as a separate slide bag, but all slides from the same TCGA case are kept in the same split. For cases with multiple WSIs, we average slide-level risk scores to obtain one case-level risk score. Validation model selection uses a case-level C-index. If a validation split has no comparable pairs, validation loss is used as the fallback selection signal. The reported TCGA results use the test case-level C-index.

\paragraph{Splits and leakage control.}
All main WSI experiments use five global seeds from $42$ to $46$. For each seed, we generate an approximate $70\%$, $10\%$, and $20\%$ split for training, validation, and testing. The same split is reused by all compared methods under the same dataset and seed, which gives a paired comparison across attention modules. The split unit is chosen to avoid leakage when multiple slides may come from the same subject. TCGA survival is split by case. CAMELYON17 is split by patient and stratified by center. BRACS is split by patient using a patient-level label vector, since a single patient may contribute slides from multiple diagnostic categories. CAMELYON16, KGH, and PANDA use slide-level split units because the available experimental inventory does not provide the separate patient grouping used by the trainer. These splits are stratified by class label for CAMELYON16 and KGH, and by data provider and ISUP grade for PANDA. The architecture choice ablation for PANDA \texttt{vit4} uses the same global-seed slide split as the corresponding PANDA classification package. ADP uses the official Release1 patch-row train, validation, and test split for every method; seeds $42$ to $46$ control model initialization and training stochasticity rather than ADP split membership.

\paragraph{WSI token handling.}
All WSI loaders use one WSI bag per mini-batch, and the effective batch size is obtained via gradient accumulation. We do not apply stochastic patch sampling during training or evaluation. For each WSI, all extracted tissue patch tokens are used unless a slide exceeds the deterministic memory cap for its dataset. The caps are $16{,}384$ tokens for CAMELYON16, $49{,}152$ tokens for CAMELYON17 and TCGA, and $32{,}768$ tokens for KGH and BRACS. If a cap is triggered, tokens are deterministically selected in coordinate order with uniform spacing. This ensures that all methods and splits use the same reproducible subset for that slide. PANDA, including the dense \texttt{vit4} ablation, is run at native slide length without a cap. ADP is not a WSI bag experiment: each image contributes the full $17 \times 17$ ViT token grid.

\paragraph{Optimization.}
We train PANDA for $20$ epochs because it is the largest and most heterogeneous classification dataset. The other slide-level classification tasks are trained for $15$ epochs to avoid unnecessary overfitting on smaller cohorts. Slide-level classification runs use gradient accumulation of $16$. We use AdamW~\cite{AdamW} with $\beta_1=0.9$, $\beta_2=0.999$, $5\%$ linear warmup, cosine decay, learning rate $2\times10^{-4}$, and weight decay $0.05$. Gate and projection parameters that encode additive coefficients or biases are excluded from weight decay. The ADP experiment is trained for $20$ epochs with batch size $256$, AdamW, learning rate $5\times10^{-4}$, weight decay $0.05$, $5\%$ warmup, cosine decay, and gradient clipping at norm $1.0$. Survival tasks are trained for $20$ epochs with AdamW, learning rate $1\times10^{-4}$, weight decay $1\times10^{-5}$, gradient accumulation of $1$, $5\%$ warmup, and cosine decay.

\paragraph{Hyperparameter Protocol}
Hyperparameters are handled with fixed comparator settings. Standard NA, XSA, and Diff do not introduce specific search settings beyond the shared optimization and checkpointing protocol. Gated SRP uses one selected signed range and gate capacity for each WSI dataset or TCGA cohort, reported in Table~\ref{tab:gated-config}; the architecture choice ablation settings are reported separately in Table~\ref{tab:dense-attention-ablation-config}. This keeps the comparison focused on the attention mechanisms while making settings for each method explicit.


Table~\ref{tab:gated-config} lists the selected Gated SRP settings used in the reported WSI experiments. Here $\delta$ denotes the signed gate range, and $h$ is the hidden dimension of the token gate MLP. The selected values are fixed before final result aggregation.

\begin{table}[ht]
\centering
\caption{\textbf{Architecture choice ablation hyperparameters.} These settings apply only to the ADP/PANDA architecture choice ablation. ADP uses the raw-RGB ViT-S/16 patch classifier, and PANDA uses the dense slide-level \texttt{vit4} set-style ViT.}
\label{tab:dense-attention-ablation-config}
\begin{tabular}{lcc}
\toprule
\textbf{Dataset} & $\boldsymbol{\delta}$ & $\boldsymbol{h}$ \\
\midrule
ADP & 0.5 & 128 \\
PANDA & 2.0 & 32 \\
\bottomrule
\end{tabular}
\end{table}

\begin{table}[ht]
\centering
\caption{\textbf{Selected Gated SRP hyperparameters.} We report the selected signed gate range and gate hidden dimension for WSI classification datasets and TCGA survival cohorts. $\delta$ denotes the signed gate range $[-\delta,\delta]$, and $h$ denotes the hidden dimension of the gate MLP.}
\label{tab:gated-config}
\begin{tabular}{lcc@{\hspace{1.2cm}}lcc}
\toprule
\multicolumn{3}{c}{\textbf{Survival Analysis}} &
\multicolumn{3}{c}{\textbf{WSI Classification}} \\
\cmidrule(lr){1-3}
\cmidrule(lr){4-6}
\textbf{Cohort} & $\boldsymbol{\delta}$ & $\boldsymbol{h}$ &
\textbf{Dataset} & $\boldsymbol{\delta}$ & $\boldsymbol{h}$ \\
\midrule
TCGA-KIRC & 1.5 & 128 & CAMELYON16 & 1.0 & 16  \\
TCGA-KIRP & 2.0 & 32 & CAMELYON17 & 2.0 & 64   \\
TCGA-LUAD & 1.0 & 128 & KGH        & 1.0 & 64  \\
TCGA-STAD & 0.5 & 128 & PANDA      & 0.5 & 16  \\
TCGA-UCEC & 1.5 & 32 & BRACS      & 0.5 & 128 \\
\bottomrule
\end{tabular}
\end{table}
For the patch-encoder ablation, we also run hyperparameter search for the Gated SRP on each dataset. For UNI-v2, we use the hyperparameters from Table~\ref{tab:gated-config}. For ViT-B/16 and MedSigLIP encoders, the selected configurations are shown in Table~\ref{tab:patch-encoder-gated-config}.

\begin{table}[ht]
\centering
\caption{\textbf{Patch-encoder ablation hyperparameters.}}
\label{tab:patch-encoder-gated-config}
\begin{tabular}{lcc@{\hspace{1.0cm}}lcc}
\toprule
\multicolumn{3}{c}{\textbf{ViT-B/16}} &
\multicolumn{3}{c}{\textbf{MedSigLIP-448}} \\
\cmidrule(lr){1-3}\cmidrule(lr){4-6}
\textbf{Dataset/cohort} &
$\boldsymbol{\delta}$ &
$\boldsymbol{h}$ &
\textbf{Dataset/cohort} &
$\boldsymbol{\delta}$ &
$\boldsymbol{h}$ \\
\midrule
KGH & 1.0 & 64 &
KGH & 1.0 & 64 \\
PANDA & 0.5 & 16 &
PANDA & 0.5 & 16 \\
BRACS & 2.0 & 128 &
BRACS & 0.5 & 32 \\
TCGA-KIRP & 0.5 & 64 &
TCGA-KIRP & 2.0 & 32 \\
TCGA-LUAD & 2.0 & 32 &
TCGA-LUAD & 1.0 & 128 \\
TCGA-STAD & 0.5 & 128 &
TCGA-STAD & 0.5 & 128 \\
\bottomrule
\end{tabular}
\end{table}

\section{Pseudocode}

We present the full per-block implementation of our proposed Gated SRP, which fits in roughly twenty lines.

\begin{algorithm}[ht]
\caption{Gated SRP attention block in PyTorch-style pseudocode.
}
\label{alg:gated-srp}
\footnotesize
\begin{algorithmic}[1]
\Require \texttt{x}; projections \texttt{Wq, Wk, Wv, Wo}; neighbour
  graph \texttt{nbr\_idx, nbr\_mask}; \texttt{h\_local}; gate
  parameters \texttt{g\_tok, w\_h, b\_h, b\_lh}; range \texttt{delta}.
\Ensure \texttt{out} (replaces the standard attention-block output).
\Statex \textbf{\textit{\# Standard multi-head self-attention}}
\State \texttt{Q, K, V = split\_heads(x @ Wq, x @ Wk, x @ Wv)}
\State \texttt{A = softmax(Q @ K.transpose(-2,-1) / sqrt(d))}
\State \texttt{Y = A @ V} \Comment{SA output}
\Statex \textbf{\textit{\# === Gated SRP ===}}
\State \texttt{V\_nbr = gather(V.detach(), nbr\_idx)}
  \Comment{valid neighbours only define the geometric reference}
\State \texttt{mask = nbr\_mask[:,None,:,:,None].float()}
\State \texttt{m = mask.sum(-2).clamp\_min(1.0)}
\State \texttt{r = (V\_nbr * mask).sum(-2) / m}
  \Comment{Eq. (6)}
\State \texttt{r\_hat = r / (r.norm(-1, keepdim=True) + eps)}
\State \texttt{c = (Y * r\_hat).sum(-1, keepdim=True) * r\_hat}
  \Comment{Eq. (4)}
\State \texttt{m\_tok = nbr\_mask.sum(-1).float()}
\State \texttt{has\_nbr = (m\_tok > 0).float()[:,None,:,None]}
\State \texttt{d\_tok = stack([h\_local, m\_tok/8, log1p(m\_tok)], -1).detach()}
  \Comment{Eq. (9)}
\State \texttt{cos\_yr = (Y * r\_hat).sum(-1)}
\State \texttt{d\_head = stack([cos\_yr, cos\_yr.abs(),}
\Statex \hspace{3.2em}\texttt{log1p(Y.norm(-1))], -1).detach()}
  \Comment{Eq. (10)}
\State \texttt{logit = g\_tok(d\_tok)[:,None,:,None]}
\Statex \hspace{3.2em}\texttt{+ einsum("bhnt,ht->bhn", d\_head, w\_h)[...,None]}
\Statex \hspace{3.2em}\texttt{+ b\_h.view(1,H,1,1) + b\_lh.view(1,H,1,1)}
  \Comment{Eq. (8)}
\State \texttt{beta = delta * tanh(logit) * has\_nbr}
  \Comment{Eq. (7)}
\State \texttt{Z = Y - beta * c}
  \Comment{replaces \texttt{Y}; Eq. (5)}
\Statex \textbf{\textit{\# === end Gated SRP ===}}
\State \texttt{out = x + Wo @ merge\_heads(Z)}
  \Comment{standard residual + output projection}
\end{algorithmic}
\end{algorithm}

\section{Standard Deviations of Main Results}
\label{sec:supp-main-std}


\begin{table}[ht]
\centering
\caption{\textbf{Survival standard deviations.}
We report the C-index standard deviation over five seeds.
Bold marks the lowest value for each cohort.}
\label{tab:survival-main-std}
{\scriptsize
\setlength{\tabcolsep}{2.5pt}
\renewcommand{\arraystretch}{1.04}
\begin{tabular}{@{}lccccc@{}}
\toprule
\textbf{Method} & \textbf{KIRC} & \textbf{KIRP} & \textbf{LUAD} & \textbf{STAD} & \textbf{UCEC} \\
\midrule
NA~\cite{transmil,nystrom}
& 0.0516 & 0.1244 & \textbf{0.0424} & 0.0672 & 0.0818 \\
XSA~\cite{XSA-paper}
& 0.0456 & 0.1191 & 0.0642 & 0.0799 & \textbf{0.0528} \\
Diff~\cite{diff}
& \textbf{0.0394} & 0.0884 & 0.0485 & 0.0638 & 0.0869 \\
\textbf{Gated SRP}
& 0.0634 & \textbf{0.0805} & 0.0492 & \textbf{0.0542} & 0.0844 \\
\bottomrule
\end{tabular}}
\end{table}

The standard deviations observed in the main survival and classification experiments are reported in Tables~\ref{tab:survival-main-std} and~\ref{tab:main-classification-std}. We conduct experiments using five global seeds to control data splitting, model initialization, and training stochasticity. As a result, the reported standard deviations capture the full evaluation protocol, not just variation from training on a fixed test set. In survival analysis, each seed alters the held-out case set and its event and censoring distribution, which directly impacts C-index evaluation. Table~\ref{tab:survival-main-std} shows that survival metric variation is present across all attention mechanisms, not only Gated SRP. Gated SRP achieves the lowest standard deviation on KIRP and STAD, while other methods are lower on KIRC, LUAD, and UCEC. For classification, standard deviations are typically smaller for AUC and larger for thresholded metrics such as F1 and accuracy, particularly in smaller or imbalanced multi-class datasets. Gated SRP reduces standard deviation compared to NA in 9 out of 16 classification metrics. In conclusion, the observed mean performance improvements of Gated SRP are not associated with increased training instability.

\begin{table*}[ht]
\centering
\caption{\textbf{Classification standard deviations.} The standard deviations for the main classification results. The lowest standard deviation for each metric is \textbf{bolded}.}
\label{tab:main-classification-std}
{\scriptsize
\setlength{\tabcolsep}{2.1pt}
\renewcommand{\arraystretch}{1.04}
\resizebox{\textwidth}{!}{%
\begin{tabular}{l*{16}{c}}
\toprule
\textbf{Method} & \multicolumn{3}{c}{\textbf{CAM16}} & \multicolumn{3}{c}{\textbf{CAM17}} & \multicolumn{3}{c}{\textbf{KGH}} & \multicolumn{4}{c}{\textbf{PANDA}} & \multicolumn{3}{c}{\textbf{BRACS}} \\
\cmidrule(lr){2-4}\cmidrule(lr){5-7}\cmidrule(lr){8-10}\cmidrule(lr){11-14}\cmidrule(lr){15-17}
& \textit{F1} & \textit{Acc} & \textit{AUC} & \textit{F1} & \textit{Acc} & \textit{AUC} & \textit{F1} & \textit{Acc} & \textit{AUC} & $\mathit{\kappa_q}$ & \textit{F1} & \textit{Acc} & \textit{AUC} & \textit{F1} & \textit{Acc} & \textit{AUC} \\
\midrule
NA~\cite{transmil, nystrom}& 0.0218 & 0.0211 & 0.0035 & \textbf{0.0449} & 0.0205 & 0.0305 & 0.0487 & 0.0423 & \textbf{0.0080} & 0.0097 & 0.0141 & 0.0111 & 0.0154 & 0.0659 & 0.0504 & \textbf{0.0180} \\
XSA~\cite{XSA-paper} & 0.0172 & 0.0166 & 0.0036 & 0.0572 & 0.0239 & 0.0310 & 0.0283 & 0.0284 & 0.0116 & \textbf{0.0036} & 0.0162 & 0.0143 & 0.0077 & \textbf{0.0514} & \textbf{0.0496} & 0.0208 \\
Diff~\cite{diff} & \textbf{0.0106} & \textbf{0.0101} & 0.0064 & 0.0516 & 0.0260 & \textbf{0.0154} & \textbf{0.0242} & \textbf{0.0239} & 0.0089 & 0.0055 & 0.0134 & \textbf{0.0075} & \textbf{0.0038} & 0.0853 & 0.0544 & 0.0357 \\
\midrule
\textbf{Gated SRP} & 0.0233 & 0.0227 & \textbf{0.0014} & 0.0549 & \textbf{0.0170} & 0.0291 & 0.0336 & 0.0331 & 0.0084 & 0.0057 & \textbf{0.0132} & 0.0101 & 0.0116 & 0.0877 & 0.0694 & 0.0265 \\
\bottomrule
\end{tabular}}}
\end{table*}

\section{Additional Gate Trajectories}
\label{sec:supp-beta-trajectory}

\begin{figure*}[ht]
\centering
\includegraphics[width=0.8\textwidth]{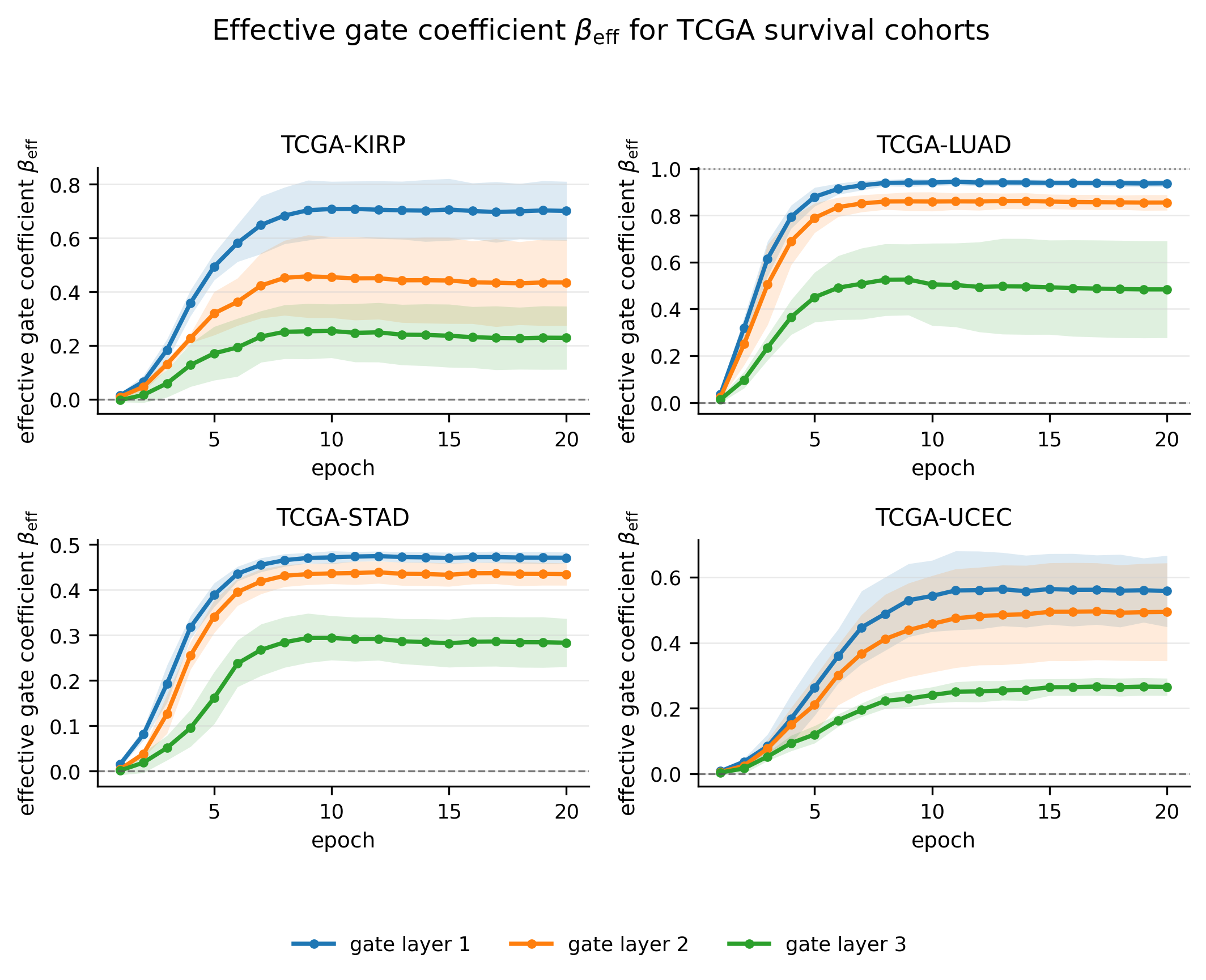}
\caption{\textbf{Effective gate coefficient trajectories for TCGA survival cohorts.} The additional TCGA survival cohorts. Curves and shaded regions follow the same five-seed mean and $\pm1$ standard-deviation convention.}
\label{fig:supp-beta-trajectory-survival}
\end{figure*}

It is important to distinguish the selected range parameter and the realized gate coefficients in Gated SRP. The effective correction coefficient is defined as $\beta=\delta\tanh(s)$, where $s$ is the gate logit. The parameter $\delta$ sets both the maximum possible value of $\beta$ and the gradient scale of the gate output, as $\partial \beta / \partial s = \delta(1-\tanh^2(s))$. Increasing $\delta$ expands the feasible correction range and the update scale, while decreasing $\delta$ restricts both. Despite this, the final learned $\beta$ values depend on the optimized gate logits. Empirically, datasets with different $\delta$ values can yield similar ranges of realized gate coefficients. Therefore, we focus on analyzing the learned effective $\beta$ trajectories, rather than relying on $\delta$ alone to interpret the magnitude of the correction.

In the main text, we show TCGA-KIRC, CAMELYON16, and PANDA as representative examples. Here we present the gate trajectory analyses for the remaining datasets. Similarly, each curve is the mean of the seed-level epoch means over five global seeds, and the shaded region is $\pm1$ standard deviation across seeds.

The results for the remaining TCGA survival cohorts, shown in Figure~\ref{fig:supp-beta-trajectory-survival}, indicate that the learned gate coefficients are consistently positive and reach their highest values in the earliest active layer. In LUAD, the first two layers approach the projection regime, while the third layer is weaker and shows greater variability across seeds. KIRP and UCEC exhibit moderate positive corrections with broader variation, reflecting the influence of evaluation case composition on survival ranking. STAD, which uses a smaller selected range ($\delta=0.5$), has gate coefficients that approach the upper bound of this range but do not reach the projection regime. These findings suggest that reducing local common components in early layers supports more effective prognostic modeling before deeper slide-level aggregation.

\begin{figure*}[ht]
\centering
\includegraphics[width=\textwidth]{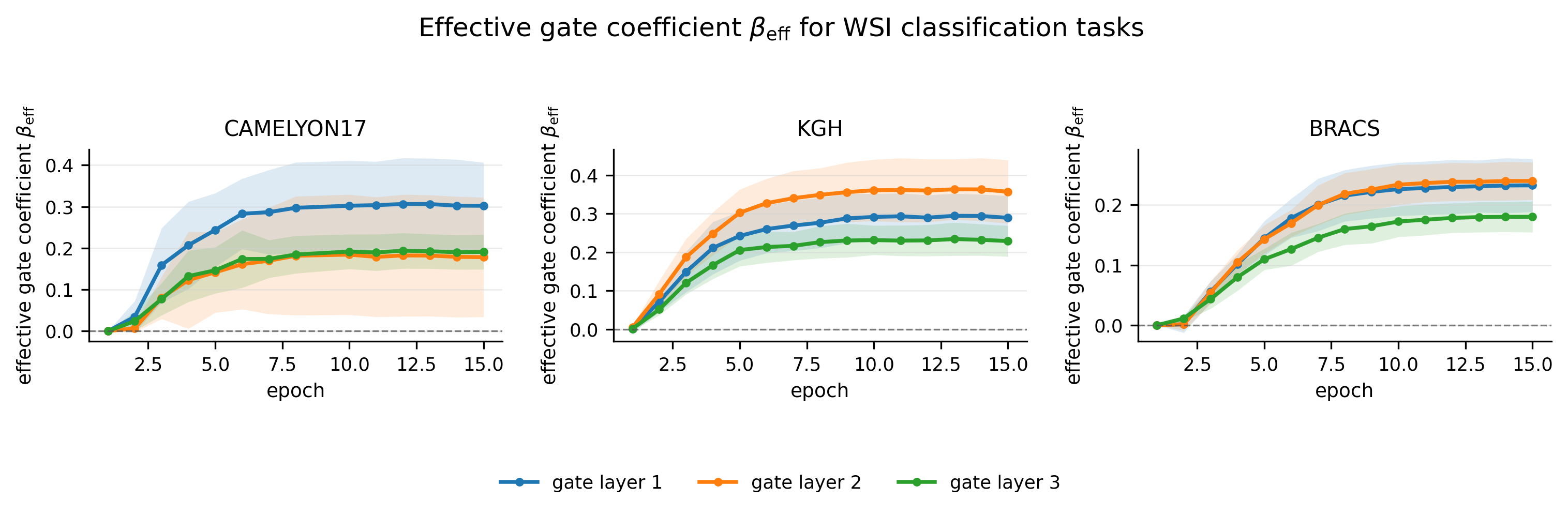}
\caption{\textbf{Effective gate coefficient trajectories for WSI classification tasks.} The additional classification datasets.}
\label{fig:supp-beta-trajectory-classification}
\end{figure*}

Figure~\ref{fig:supp-beta-trajectory-classification} presents the gate coefficient trajectories for the remaining classification datasets. The learned corrections are smaller than those in the strongest survival cohorts. In CAMELYON17, the first layer shows a moderate positive correction, while the second and third layers are lower and more variable across seeds. KGH and BRACS display modest positive gate values, with the middle layer slightly stronger than the first. This pattern aligns with classification tasks where label evidence is more directly associated with visible lesion regions, leading the model to apply more conservative corrections. In summary, Gated SRP adapts its correction magnitude by dataset and layer, ranging from near-identity updates to positive redundancy reduction or small signed corrections.

%% file: egbib.bib
@misc{XSA-paper,
      title={Exclusive Self Attention}, 
      author={Shuangfei Zhai},
      year={2026},
      eprint={2603.09078},
      archivePrefix={arXiv},
      primaryClass={cs.LG},
      url={https://arxiv.org/abs/2603.09078}, 
}

@article{vit,
  title={An Image is Worth 16x16 Words: Transformers for Image Recognition at Scale},
  author={Dosovitskiy, Alexey and Beyer, Lucas and Kolesnikov, Alexander and Weissenborn, Dirk and Zhai, Xiaohua and Unterthiner, Thomas and  Dehghani, Mostafa and Minderer, Matthias and Heigold, Georg and Gelly, Sylvain and Uszkoreit, Jakob and Houlsby, Neil},
  journal={ICLR},
  year={2021}
}

@inproceedings{diff,
 author = {Ye, Tianzhu and Dong, Li and Xia, Yuqing and Sun, Yutao and Zhu, Yi and Huang, Gao and Wei, Furu},
 booktitle = {International Conference on Learning Representations},
 editor = {Y. Yue and A. Garg and N. Peng and F. Sha and R. Yu},
 pages = {144--164},
 title = {Differential Transformer},
 url = {https://proceedings.iclr.cc/paper_files/paper/2025/file/00b67df24009747e8bbed4c2c6f9c825-Paper-Conference.pdf},
 volume = {2025},
 year = {2025}
}

@InProceedings{cait,
    author    = {Touvron, Hugo and Cord, Matthieu and Sablayrolles, Alexandre and Synnaeve, Gabriel and J\'egou, Herv\'e},
    title     = {Going Deeper With Image Transformers},
    booktitle = {Proceedings of the IEEE/CVF International Conference on Computer Vision (ICCV)},
    month     = {October},
    year      = {2021},
    pages     = {32-42}
}

@misc{specialization,
      title={Revisiting [CLS] and Patch Token Interaction in Vision Transformers}, 
      author={Alexis Marouani and Oriane Siméoni and Hervé Jégou and Piotr Bojanowski and Huy V. Vo},
      year={2026},
      eprint={2602.08626},
      archivePrefix={arXiv},
      primaryClass={cs.CV},
      url={https://arxiv.org/abs/2602.08626}, 
}

@article{cpath,
title = {Computational pathology: A survey review and the way forward},
journal = {Journal of Pathology Informatics},
volume = {15},
pages = {100357},
year = {2024},
issn = {2153-3539},
doi = {https://doi.org/10.1016/j.jpi.2023.100357},
url = {https://www.sciencedirect.com/science/article/pii/S2153353923001712},
author = {Mahdi S. Hosseini and Babak Ehteshami Bejnordi and Vincent Quoc-Huy Trinh and Lyndon Chan and Danial Hasan and Xingwen Li and Stephen Yang and Taehyo Kim and Haochen Zhang and Theodore Wu and Kajanan Chinniah and Sina Maghsoudlou and Ryan Zhang and Jiadai Zhu and Samir Khaki and Andrei Buin and Fatemeh Chaji and Ala Salehi and Bich Ngoc Nguyen and Dimitris Samaras and Konstantinos N. Plataniotis},
}

@article{titan,
  title={A multimodal whole-slide foundation model for pathology},
  author={Ding, Tong and Wagner, Sophia J and Song, Andrew H and Chen, Richard J and Lu, Ming Y and Zhang, Andrew and Vaidya, Anurag J and Jaume, Guillaume and Shaban, Muhammad and Kim, Ahrong and others},
  journal={Nature Medicine},
  pages={1--13},
  year={2025},
  publisher={Nature Publishing Group US New York}
}

@article{gigapath,
  title={A whole-slide foundation model for digital pathology from real-world data},
  author={Xu, Hanwen and Usuyama, Naoto and Bagga, Jaspreet and Zhang, Sheng and Rao, Rajesh and Naumann, Tristan and Wong, Cliff and Gero, Zelalem and González, Javier and Gu, Yu and Xu, Yanbo and Wei, Mu and Wang, Wenhui and Ma, Shuming and Wei, Furu and Yang, Jianwei and Li, Chunyuan and Gao, Jianfeng and Rosemon, Jaylen and Bower, Tucker and Lee, Soohee and Weerasinghe, Roshanthi and Wright, Bill J. and Robicsek, Ari and Piening, Brian and Bifulco, Carlo and Wang, Sheng and Poon, Hoifung},
  journal={Nature},
  year={2024},
  publisher={Nature Publishing Group UK London}
}

@article{lu2024conch,
  title={A visual-language foundation model for computational pathology},
  author={Lu, Ming Y and Chen, Bowen and Williamson, Drew FK and Chen, Richard J and Liang, Ivy and Ding, Tong and Jaume, Guillaume and Odintsov, Igor and Le, Long Phi and Gerber, Georg and others},
  journal={Nature Medicine},
  pages={863–-874},
  volume={30},
  year={2024},
  publisher={Nature Publishing Group}
}

@Article{UNI,
  title={Towards a General-Purpose Foundation Model for Computational Pathology},
  author={Chen, Richard J and Ding, Tong and Lu, Ming Y and Williamson, Drew FK and Jaume, Guillaume and Chen, Bowen and Zhang, Andrew and Shao, Daniel and Song, Andrew H and Shaban, Muhammad and others},
  journal={Nature Medicine},
  publisher={Nature Publishing Group},
  year={2024}
}

@inproceedings{transmil,
 author = {Shao, Zhuchen and Bian, Hao and Chen, Yang and Wang, Yifeng and Zhang, Jian and Ji, Xiangyang and zhang, yongbing},
 booktitle = {Advances in Neural Information Processing Systems},
 editor = {M. Ranzato and A. Beygelzimer and Y. Dauphin and P.S. Liang and J. Wortman Vaughan},
 pages = {2136--2147},
 publisher = {Curran Associates, Inc.},
 title = {TransMIL: Transformer based Correlated Multiple Instance Learning for Whole Slide Image Classification},
 url = {https://proceedings.neurips.cc/paper_files/paper/2021/file/10c272d06794d3e5785d5e7c5356e9ff-Paper.pdf},
 volume = {34},
 year = {2021}
}

@InProceedings{hipt,
    author    = {Chen, Richard J. and Chen, Chengkuan and Li, Yicong and Chen, Tiffany Y. and Trister, Andrew D. and Krishnan, Rahul G. and Mahmood, Faisal},
    title     = {Scaling Vision Transformers to Gigapixel Images via Hierarchical Self-Supervised Learning},
    booktitle = {Proceedings of the IEEE/CVF Conference on Computer Vision and Pattern Recognition (CVPR)},
    month     = {June},
    year      = {2022},
    pages     = {16144-16155}
}

@inproceedings{transformer,
 author = {Vaswani, Ashish and Shazeer, Noam and Parmar, Niki and Uszkoreit, Jakob and Jones, Llion and Gomez, Aidan N and Kaiser, \L ukasz and Polosukhin, Illia},
 booktitle = {Advances in Neural Information Processing Systems},
 editor = {I. Guyon and U. Von Luxburg and S. Bengio and H. Wallach and R. Fergus and S. Vishwanathan and R. Garnett},
 pages = {},
 publisher = {Curran Associates, Inc.},
 title = {Attention is All you Need},
 url = {https://proceedings.neurips.cc/paper_files/paper/2017/file/3f5ee243547dee91fbd053c1c4a845aa-Paper.pdf},
 volume = {30},
 year = {2017}
}

@article{adpv2,
title = {ADPv2: A hierarchical histological tissue type-annotated dataset for potential biomarker discovery of colorectal disease},
journal = {Journal of Pathology Informatics},
volume = {20},
pages = {100537},
year = {2026},
issn = {2153-3539},
doi = {https://doi.org/10.1016/j.jpi.2025.100537},
url = {https://www.sciencedirect.com/science/article/pii/S2153353925001233},
author = {Zhiyuan Yang and Kai Li and Sophia Ghamoshi Ramandi and Patricia Brassard and Abdelhakim Khellaf and Vincent Quoc-Huy Trinh and Jennifer Zhang and Lina Chen and Corwyn Rowsell and Sonal Varma and Kostas Plataniotis and Mahdi S. Hosseini},
}

@InProceedings{adp,
author = {Hosseini, Mahdi S. and Chan, Lyndon and Tse, Gabriel and Tang, Michael and Deng, Jun and Norouzi, Sajad and Rowsell, Corwyn and Plataniotis, Konstantinos N. and Damaskinos, Savvas},
title = {Atlas of Digital Pathology: A Generalized Hierarchical Histological Tissue Type-Annotated Database for Deep Learning},
booktitle = {Proceedings of the IEEE/CVF Conference on Computer Vision and Pattern Recognition (CVPR)},
month = {June},
year = {2019}
}

@article{cam16,
    author = {Ehteshami Bejnordi, Babak and Veta, Mitko and Johannes van Diest, Paul and van Ginneken, Bram and Karssemeijer, Nico and Litjens, Geert and van der Laak, Jeroen A. W. M. and and the CAMELYON16 Consortium},
    title = {Diagnostic Assessment of Deep Learning Algorithms for Detection of Lymph Node Metastases in Women With Breast Cancer},
    journal = {JAMA},
    volume = {318},
    number = {22},
    pages = {2199-2210},
    year = {2017},
    month = {12},
    issn = {0098-7484},
    doi = {10.1001/jama.2017.14585},
    url = {https://doi.org/10.1001/jama.2017.14585},
    eprint = {https://jamanetwork.com/journals/jama/articlepdf/2665774/jama_ehteshami_bejnordi_2017_oi_170113.pdf},
}

@ARTICLE{cam17,
  author={Bándi, Péter and Geessink, Oscar and Manson, Quirine and Van Dijk, Marcory and Balkenhol, Maschenka and Hermsen, Meyke and Ehteshami Bejnordi, Babak and Lee, Byungjae and Paeng, Kyunghyun and Zhong, Aoxiao and Li, Quanzheng and Zanjani, Farhad Ghazvinian and Zinger, Svitlana and Fukuta, Keisuke and Komura, Daisuke and Ovtcharov, Vlado and Cheng, Shenghua and Zeng, Shaoqun and Thagaard, Jeppe and Dahl, Anders B. and Lin, Huangjing and Chen, Hao and Jacobsson, Ludwig and Hedlund, Martin and Çetin, Melih and Halıcı, Eren and Jackson, Hunter and Chen, Richard and Both, Fabian and Franke, Jörg and Küsters-Vandevelde, Heidi and Vreuls, Willem and Bult, Peter and van Ginneken, Bram and van der Laak, Jeroen and Litjens, Geert},
  journal={IEEE Transactions on Medical Imaging}, 
  title={From Detection of Individual Metastases to Classification of Lymph Node Status at the Patient Level: The CAMELYON17 Challenge}, 
  year={2019},
  volume={38},
  number={2},
  pages={550-560},
  doi={10.1109/TMI.2018.2867350}}

@article{bracs,
    author = {Brancati, Nadia and Anniciello, Anna Maria and Pati, Pushpak and Riccio, Daniel and Scognamiglio, Giosuè and Jaume, Guillaume and De Pietro, Giuseppe and Di Bonito, Maurizio and Foncubierta, Antonio and Botti, Gerardo and Gabrani, Maria and Feroce, Florinda and Frucci, Maria},
    title = {BRACS: A Dataset for BReAst Carcinoma Subtyping in H\&amp;E Histology Images},
    journal = {Database},
    volume = {2022},
    pages = {baac093},
    year = {2022},
    month = {01},
    issn = {1758-0463},
    doi = {10.1093/database/baac093},
    url = {https://doi.org/10.1093/database/baac093},
    eprint = {https://academic.oup.com/database/article-pdf/doi/10.1093/database/baac093/46880223/baac093.pdf},
}

@Article{panda,
author={Bulten, Wouter
and Kartasalo, Kimmo
and Chen, Po-Hsuan Cameron
and Str{\"o}m, Peter
and Pinckaers, Hans
and Nagpal, Kunal
and Cai, Yuannan
and Steiner, David F.
and van Boven, Hester
and Vink, Robert
and Hulsbergen-van de Kaa, Christina
and van der Laak, Jeroen
and Amin, Mahul B.
and Evans, Andrew J.
and van der Kwast, Theodorus
and Allan, Robert
and Humphrey, Peter A.
and Gr{\"o}nberg, Henrik
and Samaratunga, Hemamali
and Delahunt, Brett
and Tsuzuki, Toyonori
and H{\"a}kkinen, Tomi
and Egevad, Lars
and Demkin, Maggie
and Dane, Sohier
and Tan, Fraser
and Valkonen, Masi
and Corrado, Greg S.
and Peng, Lily
and Mermel, Craig H.
and Ruusuvuori, Pekka
and Litjens, Geert
and Eklund, Martin
and Brilhante, Am{\'e}rico
and {\c{C}}ak{\i}r, Asl{\i}
and Farr{\'e}, Xavier
and Geronatsiou, Katerina
and Molini{\'e}, Vincent
and Pereira, Guilherme
and Roy, Paromita
and Saile, G{\"u}nter
and Salles, Paulo G. O.
and Schaafsma, Ewout
and Tschui, Jo{\"e}lle
and Billoch-Lima, Jorge
and Pereira, Em{\'i}io M.
and Zhou, Ming
and He, Shujun
and Song, Sejun
and Sun, Qing
and Yoshihara, Hiroshi
and Yamaguchi, Taiki
and Ono, Kosaku
and Shen, Tao
and Ji, Jianyi
and Roussel, Arnaud
and Zhou, Kairong
and Chai, Tianrui
and Weng, Nina
and Grechka, Dmitry
and Shugaev, Maxim V.
and Kiminya, Raphael
and Kovalev, Vassili
and Voynov, Dmitry
and Malyshev, Valery
and Lapo, Elizabeth
and Campos, Manuel
and Ota, Noriaki
and Yamaoka, Shinsuke
and Fujimoto, Yusuke
and Yoshioka, Kentaro
and Juvonen, Joni
and Tukiainen, Mikko
and Karlsson, Antti
and Guo, Rui
and Hsieh, Chia-Lun
and Zubarev, Igor
and Bukhar, Habib S. T.
and Li, Wenyuan
and Li, Jiayun
and Speier, William
and Arnold, Corey
and Kim, Kyungdoc
and Bae, Byeonguk
and Kim, Yeong Won
and Lee, Hong-Seok
and Park, Jeonghyuk
and consortium, the PANDA challenge},
title={Artificial intelligence for diagnosis and Gleason grading of prostate cancer: the PANDA challenge},
journal={Nature Medicine},
year={2022},
month={Jan},
day={01},
volume={28},
number={1},
pages={154-163},
issn={1546-170X},
doi={10.1038/s41591-021-01620-2},
url={https://doi.org/10.1038/s41591-021-01620-2}
}

@misc{NCI_TCGA,
  author       = {{National Cancer Institute}},
  title        = {{The Cancer Genome Atlas Program (TCGA)}},
  howpublished = {\url{https://www.cancer.gov/ccg/research/genome-sequencing/tcga}},
  note         = {Accessed: 2026-04-28}
}

@article{atlaspatch,
  title   = {AtlasPatch: Efficient Tissue Detection and High-throughput Patch Extraction for Computational Pathology at Scale},
  author  = {Alagha, Ahmed and Leclerc, Christopher and Kotp, Yousef and Metwally, Omar and Moras, Calvin and Rentopoulos, Peter and Rostami, Ghodsiyeh and Nguyen, Bich Ngoc and Baig, Jumanah and Khellaf, Abdelhakim and Trinh, Vincent Quoc-Huy and Mizouni, Rabeb and Otrok, Hadi and Bentahar, Jamal and Hosseini, Mahdi S.},
  journal = {arXiv preprint arXiv:2602.03998},
  year    = {2026}
}

@inproceedings{
AdamW,
title={Decoupled Weight Decay Regularization},
author={Ilya Loshchilov and Frank Hutter},
booktitle={International Conference on Learning Representations},
year={2019},
url={https://openreview.net/forum?id=Bkg6RiCqY7},
}

@InProceedings{xavier,
  title = 	 {Understanding the difficulty of training deep feedforward neural networks},
  author = 	 {Glorot, Xavier and Bengio, Yoshua},
  booktitle = 	 {Proceedings of the Thirteenth International Conference on Artificial Intelligence and Statistics},
  pages = 	 {249--256},
  year = 	 {2010},
  editor = 	 {Teh, Yee Whye and Titterington, Mike},
  volume = 	 {9},
  series = 	 {Proceedings of Machine Learning Research},
  address = 	 {Chia Laguna Resort, Sardinia, Italy},
  month = 	 {13--15 May},
  publisher =    {PMLR},
  url = 	 {https://proceedings.mlr.press/v9/glorot10a.html}
}

@inproceedings{kaiming,
author = {He, Kaiming and Zhang, Xiangyu and Ren, Shaoqing and Sun, Jian},
title = {Delving Deep into Rectifiers: Surpassing Human-Level Performance on ImageNet Classification},
year = {2015},
isbn = {9781467383912},
publisher = {IEEE Computer Society},
address = {USA},
url = {https://doi.org/10.1109/ICCV.2015.123},
doi = {10.1109/ICCV.2015.123},
booktitle = {Proceedings of the 2015 IEEE International Conference on Computer Vision (ICCV)},
pages = {1026–1034},
numpages = {9},
series = {ICCV '15}
}

@article{sellergren2025medgemma,
  title={MedGemma Technical Report},
  author={Sellergren, Andrew and Kazemzadeh, Sahar and Jaroensri, Tiam and Kiraly, Atilla and Traverse, Madeleine and Kohlberger, Timo and Xu, Shawn and Jamil, Fayaz and Hughes, Cían and Lau, Charles and others},
  journal={arXiv preprint arXiv:2507.05201},
  year={2025}
}

@misc{differential-gated-attention,
      title={Differential Gated Self-Attention}, 
      author={Elpiniki Maria Lygizou and Mónika Farsang and Radu Grosu},
      year={2025},
      eprint={2505.24054},
      archivePrefix={arXiv},
      primaryClass={cs.LG},
      url={https://arxiv.org/abs/2505.24054}, 
}

@misc{cog-attention,
      title={More Expressive Attention with Negative Weights}, 
      author={Ang Lv and Ruobing Xie and Shuaipeng Li and Jiayi Liao and Xingwu Sun and Zhanhui Kang and Di Wang and Rui Yan},
      year={2025},
      eprint={2411.07176},
      archivePrefix={arXiv},
      primaryClass={cs.CL},
      url={https://arxiv.org/abs/2411.07176}, 
}

@inproceedings{integral-transformer,
    title = "Integral Transformer: Denoising Attention, Not Too Much Not Too Little",
    author = "Kobyzev, Ivan  and
      Ghaddar, Abbas  and
      Hu, Dingtao  and
      Chen, Boxing",
    editor = "Christodoulopoulos, Christos  and
      Chakraborty, Tanmoy  and
      Rose, Carolyn  and
      Peng, Violet",
    booktitle = "Proceedings of the 2025 Conference on Empirical Methods in Natural Language Processing",
    month = nov,
    year = "2025",
    address = "Suzhou, China",
    publisher = "Association for Computational Linguistics",
    url = "https://aclanthology.org/2025.emnlp-main.118/",
    doi = "10.18653/v1/2025.emnlp-main.118",
    pages = "2337--2354",
    ISBN = "979-8-89176-332-6"
}

@InProceedings{convit,
  title = 	 {ConViT: Improving Vision Transformers with Soft Convolutional Inductive Biases},
  author =       {D'Ascoli, St{\'e}phane and Touvron, Hugo and Leavitt, Matthew L and Morcos, Ari S and Biroli, Giulio and Sagun, Levent},
  booktitle = 	 {Proceedings of the 38th International Conference on Machine Learning},
  pages = 	 {2286--2296},
  year = 	 {2021},
  editor = 	 {Meila, Marina and Zhang, Tong},
  volume = 	 {139},
  series = 	 {Proceedings of Machine Learning Research},
  month = 	 {18--24 Jul},
  publisher =    {PMLR},
  url = 	 {https://proceedings.mlr.press/v139/d-ascoli21a.html}
}

@InProceedings{swin,
    author    = {Liu, Ze and Lin, Yutong and Cao, Yue and Hu, Han and Wei, Yixuan and Zhang, Zheng and Lin, Stephen and Guo, Baining},
    title     = {Swin Transformer: Hierarchical Vision Transformer Using Shifted Windows},
    booktitle = {Proceedings of the IEEE/CVF International Conference on Computer Vision (ICCV)},
    month     = {October},
    year      = {2021},
    pages     = {10012-10022}
}

@InProceedings{neighborhood-attention,
    author    = {Hassani, Ali and Walton, Steven and Li, Jiachen and Li, Shen and Shi, Humphrey},
    title     = {Neighborhood Attention Transformer},
    booktitle = {Proceedings of the IEEE/CVF Conference on Computer Vision and Pattern Recognition (CVPR)},
    month     = {June},
    year      = {2023},
    pages     = {6185-6194}
}

@inproceedings{dynamicvit,
 author = {Rao, Yongming and Zhao, Wenliang and Liu, Benlin and Lu, Jiwen and Zhou, Jie and Hsieh, Cho-Jui},
 booktitle = {Advances in Neural Information Processing Systems},
 editor = {M. Ranzato and A. Beygelzimer and Y. Dauphin and P.S. Liang and J. Wortman Vaughan},
 pages = {13937--13949},
 publisher = {Curran Associates, Inc.},
 title = {DynamicViT: Efficient Vision Transformers with Dynamic Token Sparsification},
 url = {https://proceedings.neurips.cc/paper_files/paper/2021/file/747d3443e319a22747fbb873e8b2f9f2-Paper.pdf},
 volume = {34},
 year = {2021}
}

@misc{evit,
      title={Not All Patches are What You Need: Expediting Vision Transformers via Token Reorganizations}, 
      author={Youwei Liang and Chongjian Ge and Zhan Tong and Yibing Song and Jue Wang and Pengtao Xie},
      year={2022},
      eprint={2202.07800},
      archivePrefix={arXiv},
      primaryClass={cs.CV},
      url={https://arxiv.org/abs/2202.07800}, 
}

@misc{tokenlearner,
      title={TokenLearner: What Can 8 Learned Tokens Do for Images and Videos?}, 
      author={Michael S. Ryoo and AJ Piergiovanni and Anurag Arnab and Mostafa Dehghani and Anelia Angelova},
      year={2022},
      eprint={2106.11297},
      archivePrefix={arXiv},
      primaryClass={cs.CV},
      url={https://arxiv.org/abs/2106.11297}, 
}

@misc{tome,
      title={Token Merging: Your ViT But Faster}, 
      author={Daniel Bolya and Cheng-Yang Fu and Xiaoliang Dai and Peizhao Zhang and Christoph Feichtenhofer and Judy Hoffman},
      year={2023},
      eprint={2210.09461},
      archivePrefix={arXiv},
      primaryClass={cs.CV},
      url={https://arxiv.org/abs/2210.09461}, 
}

@article{clam,
  title={Data-efficient and weakly supervised computational pathology on whole-slide images},
  author={Lu, Ming Y and Williamson, Drew FK and Chen, Tiffany Y and Chen, Richard J and Barbieri, Matteo and Mahmood, Faisal},
  journal={Nature Biomedical Engineering},
  volume={5},
  number={6},
  pages={555--570},
  year={2021},
  publisher={Nature Publishing Group}
}

@InProceedings{DSMIL,
    author    = {Li, Bin and Li, Yin and Eliceiri, Kevin W.},
    title     = {Dual-Stream Multiple Instance Learning Network for Whole Slide Image Classification With Self-Supervised Contrastive Learning},
    booktitle = {Proceedings of the IEEE/CVF Conference on Computer Vision and Pattern Recognition (CVPR)},
    month     = {June},
    year      = {2021},
    pages     = {14318-14328}
}

@InProceedings{patch-gcn,
author="Chen, Richard J.
and Lu, Ming Y.
and Shaban, Muhammad
and Chen, Chengkuan
and Chen, Tiffany Y.
and Williamson, Drew F. K.
and Mahmood, Faisal",
editor="de Bruijne, Marleen
and Cattin, Philippe C.
and Cotin, St{\'e}phane
and Padoy, Nicolas
and Speidel, Stefanie
and Zheng, Yefeng
and Essert, Caroline",
title="Whole Slide Images are 2D Point Clouds: Context-Aware Survival Prediction Using Patch-Based Graph Convolutional Networks",
booktitle="Medical Image Computing and Computer Assisted Intervention -- MICCAI 2021",
year="2021",
publisher="Springer International Publishing",
address="Cham",
pages="339--349",
isbn="978-3-030-87237-3"
}

@article{splice,
title = {Sequential Patching Lattice for Image Classification and Enquiry: Streamlining Digital Pathology Image Processing},
journal = {The American Journal of Pathology},
volume = {194},
number = {10},
pages = {1898-1912},
year = {2024},
issn = {0002-9440},
doi = {https://doi.org/10.1016/j.ajpath.2024.06.007},
url = {https://www.sciencedirect.com/science/article/pii/S0002944024002384},
author = {Areej Alsaafin and Peyman Nejat and Abubakr Shafique and Jibran Khan and Saghir Alfasly and Ghazal Alabtah and Hamid R. Tizhoosh}
}

@InProceedings{focus,
    author    = {Guo, Zhengrui and Xiong, Conghao and Ma, Jiabo and Sun, Qichen and Feng, Lishuang and Wang, Jinzhuo and Chen, Hao},
    title     = {FOCUS: Knowledge-enhanced Adaptive Visual Compression for Few-shot Whole Slide Image Classification},
    booktitle = {Proceedings of the IEEE/CVF Conference on Computer Vision and Pattern Recognition (CVPR)},
    month     = {June},
    year      = {2025},
    pages     = {15590-15600}
}

@inproceedings{pathvq,
 author = {Li, Honglin and Shui, Zhongyi and Zhang, Yunlong and Zhu, Chenglu and Yang, Lin},
 booktitle = {Advances in Neural Information Processing Systems},
 editor = {D. Belgrave and C. Zhang and H. Lin and R. Pascanu and P. Koniusz and M. Ghassemi and N. Chen},
 pages = {62348--62375},
 publisher = {Curran Associates, Inc.},
 title = {PathVQ: Reforming Computational Pathology Foundation Model for Whole Slide Image Analysis via Vector Quantization},
 url = {https://proceedings.neurips.cc/paper_files/paper/2025/file/59f278de1619bdb6b53fd04e8e0976e0-Paper-Conference.pdf},
 volume = {38},
 year = {2025}
}

@misc{tcssa,
      title={TC-SSA: Token Compression via Semantic Slot Aggregation for Gigapixel Pathology Reasoning}, 
      author={Zhuo Chen and Shawn Young and Lijian Xu},
      year={2026},
      eprint={2603.01143},
      archivePrefix={arXiv},
      primaryClass={cs.CV},
      url={https://arxiv.org/abs/2603.01143}, 
}

@misc{moozy,
      title={MOOZY: A Patient-First Foundation Model for Computational Pathology},
      author={Yousef Kotp and Vincent Quoc-Huy Trinh and Christopher Pal and Mahdi S. Hosseini},
      year={2026},
      eprint={2603.27048},
      archivePrefix={arXiv},
      primaryClass={cs.CV},
      url={https://arxiv.org/abs/2603.27048},
}

@article{nystrom, 
    title={Nyströmformer: A Nyström-based Algorithm for Approximating Self-Attention}, 
    volume={35}, 
    url={https://ojs.aaai.org/index.php/AAAI/article/view/17664}, 
    DOI={10.1609/aaai.v35i16.17664}, 
    number={16}, 
    journal={Proceedings of the AAAI Conference on Artificial Intelligence}, 
    author={Xiong, Yunyang and Zeng, Zhanpeng and Chakraborty, Rudrasis and Tan, Mingxing and Fung, Glenn and Li, Yin and Singh, Vikas}, 
    year={2021}, 
    month={May}, 
    pages={14138–14148} 
}
